\RequirePackage{fix-cm}
\documentclass[twocolumn]{svjour3}          % twocolumn
\smartqed  % flush right qed marks, e.g. at end of proof
\usepackage{graphicx}
\usepackage{lipsum}
\usepackage{natbib}
\usepackage[ruled, linesnumbered,lined,commentsnumbered]{algorithm2e}
\usepackage{times}
\usepackage{epsfig}

\usepackage{amsmath}
\usepackage{amssymb}
\usepackage{caption}
\usepackage{url}
\usepackage{subfig}
\usepackage{float}
\usepackage{enumerate}
\usepackage{mathtools,amssymb}
\usepackage{xcolor}
\usepackage{booktabs}
\usepackage[english]{babel}
\usepackage[utf8]{inputenc}
\usepackage{diagbox}

\usepackage[noend]{algpseudocode}
\usepackage{mathtools}
\DeclareUnicodeCharacter{FB01}{fi}
\begin{document}

\title{Enhanced 3D Human Pose Estimation from Videos by using Attention-Based Neural Network with Dilated Convolutions%\thanks{Grants or other notes
}

% \subtitle{Do you have a subtitle?\\ If so, write it here}


\author{
        Ruixu Liu \and
        Ju Shen \and
        He Wang \and
        Chen Chen \and
        Sen-ching Cheung \and
        Vijayan K. Asari
}


\institute{Ruixu Liu
            \at University of Dayton, Ohio, USA\\
              \email{liur05@udayton.edu}           %  \\
        %   \and
        %   Ruixu Liu \at University of Dayton, Dayton Ohio \\
        %       \email{liur05@udayton.edu}
}


\maketitle
%\twocolumn[
% \begin{@twocolumnfalse}
\begin{abstract}

The attention mechanism provides a sequential prediction framework for learning spatial models with enhanced implicit temporal consistency. In this work, we show a systematic design (from 2D to 3D) for how conventional networks and other forms of constraints can be incorporated into the attention framework for learning long-range dependencies for the task of pose estimation. The contribution of this paper is to provide a systematic approach for designing and training of attention-based models for the end-to-end pose estimation, with the flexibility and scalability of arbitrary video sequences as input. We achieve this by adapting temporal receptive field via a multi-scale structure of dilated convolutions.  Besides, the proposed architecture can be easily adapted to a causal model enabling real-time performance. Any off-the-shelf 2D pose estimation systems, e.g. Mocap libraries, can be easily integrated in an ad-hoc fashion. Our method achieves the state-of-the-art performance and outperforms existing methods by reducing the mean per joint position error to 33.4 mm\footnote{Code is available at:\\ https://github.com/lrxjason/Attention3DHumanPose} on Human3.6M dataset. 
\end{abstract}


% \end{@twocolumnfalse}

\section{Introduction}
We introduced attention mechanism for the task of articulated 3D pose reconstruction from videos in the recent work \citep{liu2020attention}, which exploits the temporal contexts of long-range dependencies across frames. The ability to adaptively identify important frames or tensors output from each deep net layer and combine them with the advantages afforded by convolutional architectures allows for globally optimal inference through simultaneous processing .  The concept of ``attention'' is to learn optimized global alignment between pairwise data and has gained recent success in the integration with deep networks for processing mono/multi-modal data, such as text-to-speech matching \citep{Chorowski2015}, neural machine translation \citep{Bahdanau2015} and 2D human pose estimation \citep{chu2017multi}.  In this paper, we extend our original attention model further by integrating it with deep networks in both 2D and 3D domain, leading to improved estimation while preserving natural temporal coherence in videos.

Articulated 3D human pose estimation from unconstrained single images or videos is considered as an ill-posed problem due to the nonlinearity of human dynamics, occlusions, and the high-dimensional variability introduced in the wild. Traditional approaches such as multi-view capture \citep{Amin13}, marker based systems  \citep{Mandery2015} and multi-modal sensing \citep{Palmero2016} require a laborious setup process and are not practical for applications in the less controlled environment.  Recent efforts of using deep architectures have significantly advanced the state-of-the-art in 3D pose reasoning \citep{Toshev2014, Neverova2014}.  The end-to-end learning process alleviates the need of using tailor-made features or spatial constraints, thereby minimizing the characteristic errors such as double-counting image evidence \citep{Ferrari2009}. 
  While vast and powerful deep models on 3D pose prediction are emerging (from convolutional neural network (CNN) \citep{Pavlakos2017, Tekin2016, Li2015} to generative adversarial networks (GAN) \citep{Yang2018,  Chen2019}), many of these approaches focus on a single image inference, which is inclined to jittery motion or inexact body configuration. To resolve this, temporal information is taken into account for better motion consistency. Existing works can be generally classified into two categories: \emph{direct 3D estimation} and \emph{2D-to-3D estimation} \citep{Zhou2016, Chen2016}. The former explores the possibility of jointly extracting both 2D and 3D poses in a holistic manner \citep{Pavlakos2017, Varol2017}; while the latter decouples the estimation into two steps: 2D body part detection and 3D correspondence inference \citep{Chen2017, Bogo2016, Zhou2016}. We refer readers to the recent survey for more details of their respective advantages \citep{Martinez2017}. 

Our approach falls under the category of \emph{2D-to-3D estimation} with three key contributions:
\begin{enumerate}
\item  Development of a systematic approach for designing and training of attention-based models for pose estimation in three levels: 2D joints attention, 3D-to-2D projection attention, and 3D pose attention.
\item  Learning of implicit dependencies in large temporal receptive fields via a multi-scale structure of dilated convolutions. 
\item  Design of a systematic architecture for the integration of the attention-based model and dilation convolutional structure to enhance 3D pose inference to facilitate performance driven animation applications.  
\end{enumerate}

\begin{figure}

\subfloat{\includegraphics[width= \linewidth]{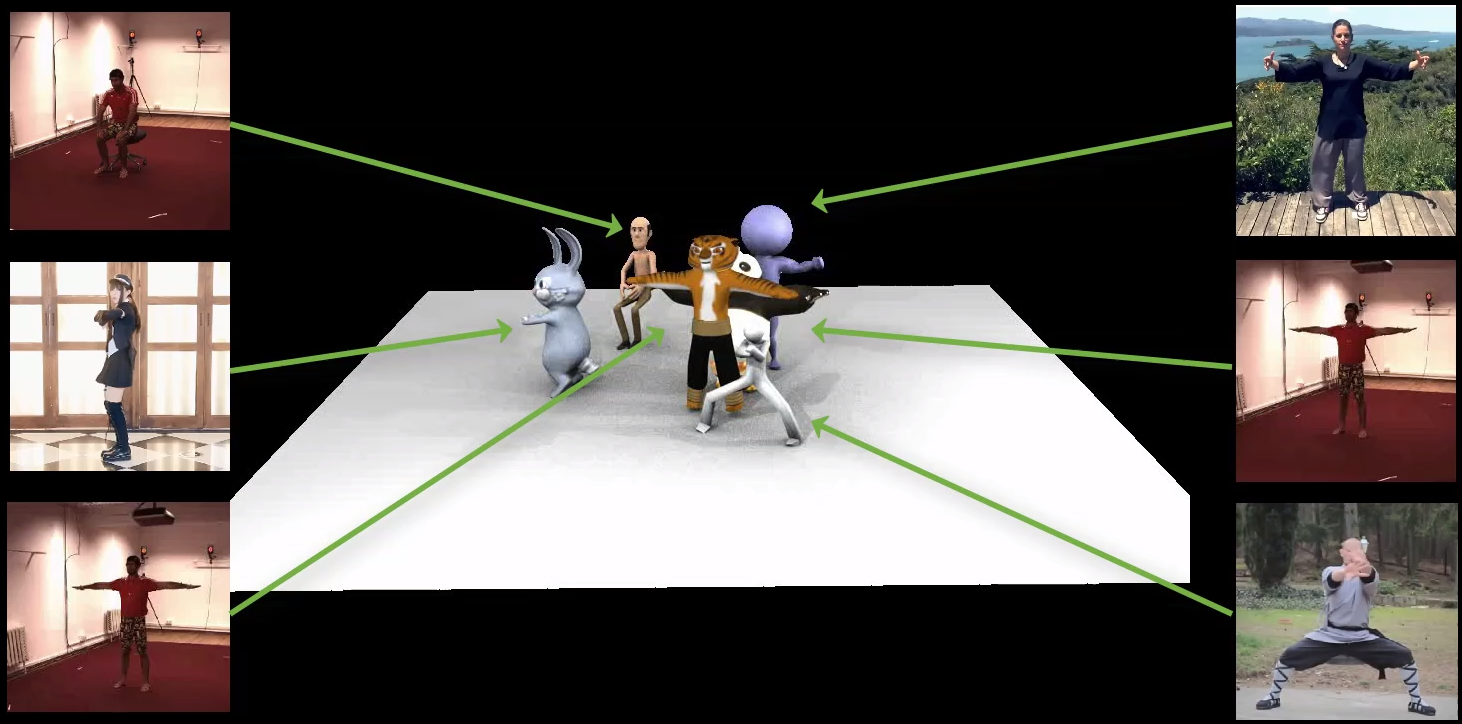}}\\
\vspace{-0.35cm}
%\subfloat{\includegraphics[width= \linewidth]{a2.png}}\\
% \vspace{-0.5cm}
\subfloat{\includegraphics[width= \linewidth]{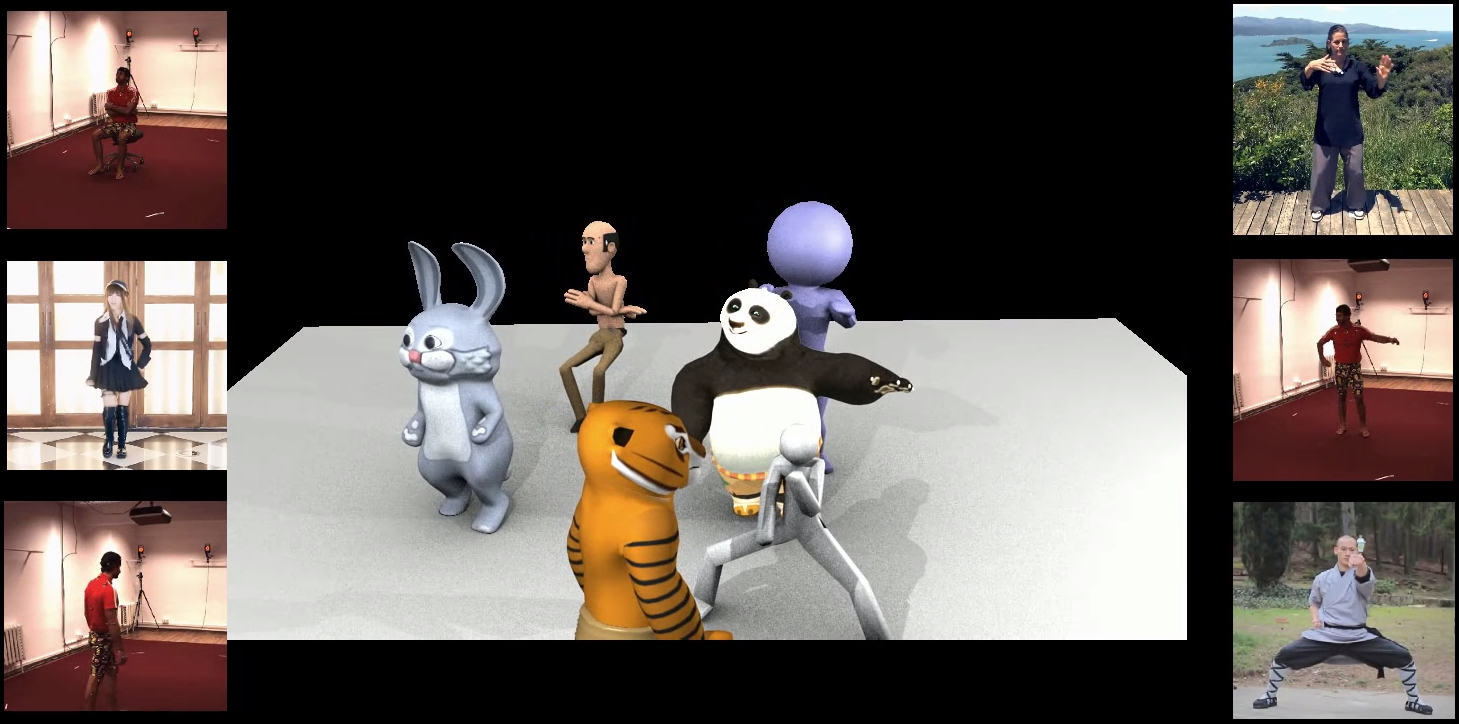}}\\
\vspace{-0.35cm}
% \subfloat{\includegraphics[width= \linewidth]{a4.png}}\\
% \vspace{-0.5cm}
\subfloat{\includegraphics[width= \linewidth]{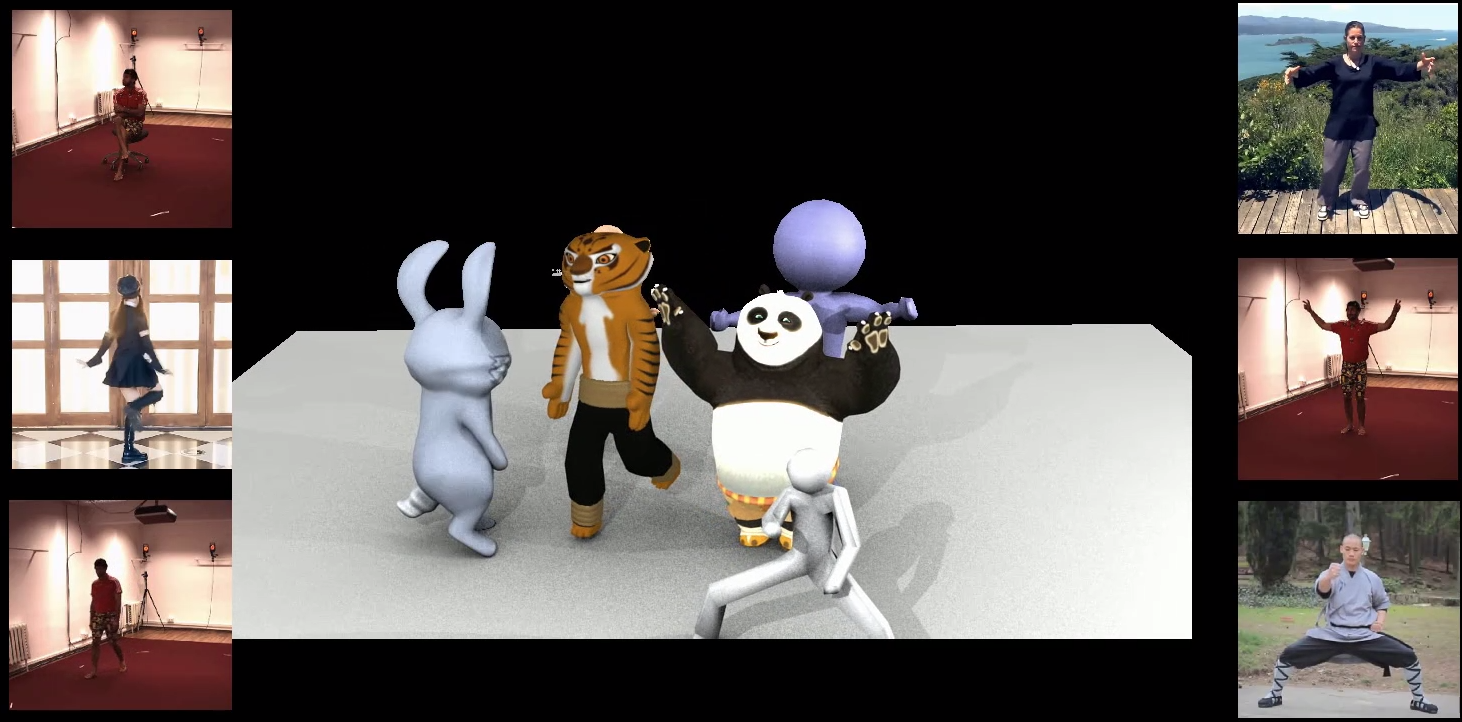}}
% \vspace{-0.5cm}
% \subfloat{\includegraphics[width= \linewidth]{ a6.png}}
\caption{An application that shows 3D avatars re-targeting from 2D video streams.}
\label{fig:all_6}
\end{figure}

Experimental evaluations show that the resulting system can reach almost the same level of estimation accuracy under both causal or non-causal conditions, making it very attractive for real-time or consumer-level applications. To date, state-of-the-art results on video-based \emph{2D-to-3D estimation} can be achieved by a semi-supervised approach \citep{Pavllo2019} or a layer normalized LSTM approach \citep{Hossain2018}. Our model can further improve the performance in both quantitative accuracy and qualitative evaluation. The simple requirement of our framework makes it well suited for interactive applications like computer games, virtual communication, and avatar animation re-targeting from videos. Given a video with continuous body movements and 3D avatars as input, we transfer the captured pose and motion from the subject video to a target character. In Fig. \ref{fig:all_6}, we show an example of how the solution can be employed in performance-based animations from videos. In this example, we create six 3D avatars with different shapes and appearances and take six different videos as input. There are not any constraints (e.g., camera intrinsic and extrinsic parameters, pose complexities, or background environment settings) about these input videos, which can be downloaded from any online sources, such as YouTube. By using the proposed technique, it enables automated body pose extraction from the video streams and applies motion re-targeting to the corresponding characters in the scene. The green arrows at the top of Fig. \ref{fig:all_6} indicates associated video for each character. The subsequent frames demonstrate the result of automatic motion transferring from the video to the 3D characters.

\section{Related Works}
\begin{figure*}
  \includegraphics[width=\linewidth]{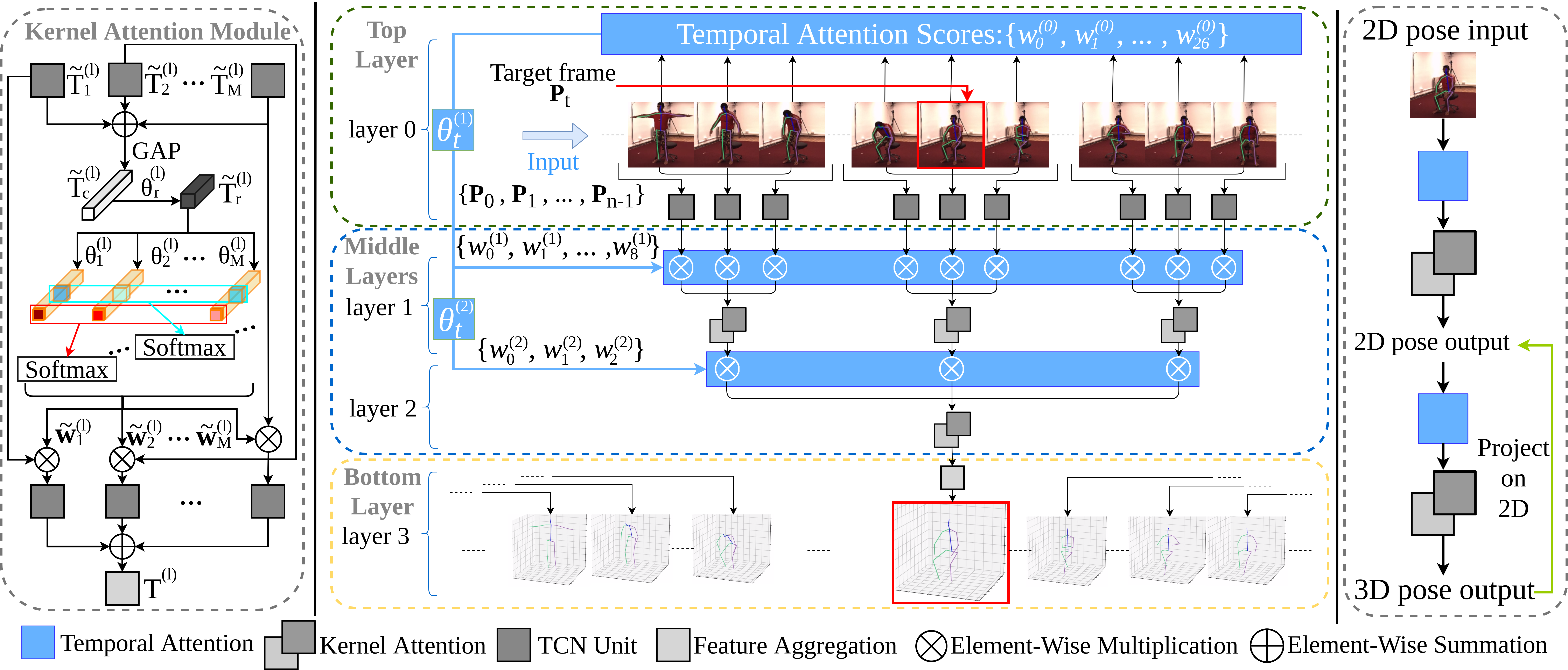}
  \caption{ An example of a 4-layer architecture for attention-based temporal convolutional neural network (ATCN). In this architecture, all the kernel sizes are 3. In practice, different layers can have different kernel sizes.}\label{fig:architecture}
\end{figure*}
Articulated pose estimation from an unconstrained video has been studied for decades. Early work relies on graphical or restrictive models to account for the high degree of freedom and dependencies among body parts, such as tree structures \citep{Andriluka2009, Yang2011, Amin13}, and pictorial structures \citep{Andriluka2009}. These methods often introduce a large number of parameters that require careful and manual tuning using techniques such as piecewise approximation. The performance of graphical model based approaches have been surpassed by convolutional neural networks (CNNs) \citep{Sarafianos2016, Pavlakos2017}, which can learn an automated representation that disentangles the dependencies among output variables without a tailor-made solver.

For the last few years, various CNN based architectures have been proposed. For example, \citep{Tekin2016} trains an auto-encoder to project human joint positions to a high dimensional space to enforce structural constraints. \citep{Park2016} estimates the 3D pose by propagating the 2D classification results to the 3D pose regressors inside a neural network \citep{Park2016}.  A  kinematic object model composing of bones and joints is introduced in \citep{Zhou20162} to guarantee the geometric validity of the estimated human body. A comprehensive list of convolutional systems can be found in the survey presented in \citep{Sarafianos2016}. 

Our contribution to this rich body of works lies in the introduction of an attention based mechanism to the body pose estimation problem. The traditional concept of “attention” is to provide an optimal matching strategy that globally aligns pairwise data from the same domain, e.g., word-to-word or phrase-to-phrase alignment in sentences \citep{Yao2013}, or across different modalities, e.g., text-to-speech \citep{Chorowski2015} and text-to-image \citep{Xu2015} in domain transformation. Prior work on attention in deep learning (DL) mostly addresses long short-term memory networks (LSTMs) \citep{Hochreiter1997} and recently it has gained popularity in training neural networks \citep{Yin2016}. Recent research indicates that certain convolutional architectures can reach state-of-the-art accuracy in audio synthesis, word-level language modeling, and machine translation \citep{oord2016wavenet, kalchbrenner2016neural, dauphin2017language}. Compared to the language modeling architecture of \citep{dauphin2017language}, temporal convolutional networks (TCNs) \citep{bai2018empirical} do not use gating mechanisms and have much longer memory. Our 3D human pose estimation and reconstruction network integrates the attention units and multi-scale dilation units to the TCN architecture.

As mentioned earlier, there are recent works that take multiple frames with 2D detection as the input for 3D prediction such as the LSTM-based method \citep{Hossain2018} and a TCN based approach with semi-supervised training \citep{Pavllo2019}. For the LSTM-based system, the frames have to be processed sequentially based on time steps, while we propose to process all of the frames in parallel for 3D pose estimation. Another objective should be that any estimation failure of one frame would not affect the other frames. In our proposed work, we also employ some similarity to the TCN-based approach as in \citep{Pavllo2019, chen2020anatomy, liu2020gast} along with the usage of a voting mechanism to select important frames for prediction. In addition, we incorporate the following three distinct features in our proposed method:

(i) Instead of making a ``hard'' decision on a subset of frames, we use a ``soft'' decision by considering all the frames. 

(ii) Along with the ``soft'' decisions to the input frames, we apply all the immediate outputs from every layer through the network, thereby expanding the scope of selection to cover both raw frames and generated features.

(iii) We use a multi-scale dilated convolution that enables us to have a broad range of frame selection without increasing the number of neural net layers.
%------------------------------------------------------------------------

\section{The Attention-based Approach} \label{sec:attention}
In this section, we present an overview of the proposed system for 3D pose estimation from a 2D video stream and show how our attention model guides the network to adaptively identify significant portion of each deep neural net layer's output resulting in an enhanced estimation.

\subsection{Network Design} 
Fig. \ref{fig:architecture} (right) depicts the overall architecture of our attention-based neural network. It takes a sequence of $n$ frames with 2D joint positions as the input and outputs the estimated 3D pose for the target frame as labeled. The framework involves two types of processing modules: the \emph{Temporal Attention} module (indicated by the long green bars) and the \emph{Kernel Attention} module (indicated by the gray squares). The kernel attention module can be further categorized as TCN Units (in dark grey color) and Feature Aggregation (in light grey color) \citep{he2016deep}. By viewing the graphical model vertically from the top, one can notice the two attention modules distribute in an interlacing pattern that a row of kernel attention modules situate right below a temporal attention module. We regard these two adjacent modules as one \emph{layer}, which has the same notion as a neural net layer. According to the functionalities, the layers can be grouped as \emph{top layer}, \emph{middle layers}, and \emph{bottom layer}. Note that the top layer only has TCN units for the kernel module, while the bottom layer only has a feature aggregation to deliver the result. It is also worth mentioning that the number of middle layers can be varied depending on the receptive field setting, which will be discussed in section \ref{ablation_study}.

\subsection{Temporal Attention}\label{units}
The goal of the temporal attention module is to provide a contribution metric for the output tensors. Each attention module produces a set of scalars, $\{\omega_{0}^{(l)}, \omega_{1}^{(l)}, \dots\}$, weighing the significance of different tensors  within a layer:

\begin{equation}
\mathbf{W}^{(l)} \otimes \mathbf{T}^{(l)} \overset{\Delta}{=} \left\{\omega_{0}^{(l)}\otimes\mathcal{T}_0^{(l)}, \dots, \omega_{\lambda_l - 1}^{(l)}\otimes\mathcal{T}_{\lambda_l - 1}^{(l)}\right\}
\label{eq:weighttensor}
\end{equation}
where $l$ and $\lambda_l$ indicate the layer index and the number of tensors output from the $l^{(th)}$ layer. We use $\mathcal{T}_u^{(l)}$ to denote the $u^{th}$ tensor output from the $l^{th}$ layer. The bold format of $\mathbf{W} \hspace{-0.5mm}\otimes\hspace{-0.5mm} \mathbf{T}$ is a compacted vector. Note for the top layer, the input to the TCN units is just the 2D joints. The choice for computing their attention scores can be flexible. A commonly used scheme is the {multilayer perceptron} strategy for optimal feature set selection \citep{Ruck1990}. Empirically, we achieve desirable result by simply computing the \emph{normalized cross-correlation ($ncc$)} that measures the positive cosine similarity between $\mathbf{P}_i$ and $\mathbf{P}_t$ on their 2D joint positions  \citep{Yoo2009}: 
\begin{equation}
\mathbf{W}^{(0)} = \left[ncc(\mathbf{P}_0, \mathbf{P}_t), \dots, ncc(\mathbf{P}_{n-1}, \mathbf{P}_t)\right]^T
\end{equation}
where $\mathbf{P}_0, \dots, \mathbf{P}_{n-1}$  are the 2D joint positions. $t$ indicates the target frame index.
The output $\mathbf{W}^{(0)}$ is forwarded to the attention matrix $\boldsymbol{\theta_t}^{(l)}$ to produce tensor weights for the subsequent layers.

\begin{equation}
\mathbf{W}^{(l)} = sig\left(\boldsymbol{\theta_t}^{(l)T}\mathbf{W}^{(l-1)}\right) \mbox{ , for } l \in [1, L-2]\label{eq: theta}
\end{equation}
where $sig(\cdot)$ is the sigmoid activation function. We require the dimension of $\boldsymbol{\theta_t}^{(l)}\in \mathcal{R}^{F'\times F}$ matching the number of output tensors between layers $l-1$ and $l$, s.t. $F' = \lambda_{l-1}$ and $F = \lambda_l$.

\subsection{Kernel Attention}
Similar to the temporal attention that determines a tensor weight distribution $\mathbf{W}^{(l)}$ within layer $l$, the kernel attention module assigns a channel weight distribution within a tensor, denoted as $\widetilde{\boldsymbol{W}}^{(l)}$. Fig. \ref{fig:architecture} (right) depicts the steps on how an updated tensor $\mathbf{T}_{final}^{(l)}$ is generated through the weight adjustment. Given an input tensor $\mathbf{T}^{(l)} \in \mathcal{R}^{C\times F}$, we generate $M$ new tensors $\widetilde{T}^{(l)}_m$ using $M$ TCN units with different dilation rates. 

These $M$ tensors are fused together through element-wise summation: $\widetilde{\mathbf{T}}^{(l)} = \sum_{m=1}^M\widetilde{T}^{(l)}_m$, which is fed into a global average pooling (GAP) layer to generate channel-wise statistics $\widetilde{\mathcal{T}}^{(l)}_c \in \mathcal{R}^{C \times 1 }$. The channel number $C$ is acquired through a TCN unit as discussed in the ablation study. The output $\widetilde{\mathcal{T}}^{(l)}_c$ is forwarded to a fully-connected layer to learn the relationship among features of different kernel sizes: $\widetilde{\mathcal{T}}^{(l)}_r = \boldsymbol{\theta_r}^{(l)}\widetilde{\mathcal{T}}^{(l)}_c$. The role of matrix $\boldsymbol{\theta_r}^{(l)} \in \mathcal{R}^{r \times C}$ is to reduce the channel dimension to $r$. Guided by the compacted feature descriptor $\widetilde{\mathcal{T}}^{(l)}_r$, $M$ vectors are generated (indicated by the yellow cuboids) through a second fully-connected layer across channels. Their kernel attention weights are computed by a softmax function:

\begin{equation}
\widetilde{\boldsymbol{W}}^{(l)} \overset{\Delta}{=} \left\{ \widetilde{W}_1^{(l)}, ..., \widetilde{W}_M^{(l)} \left|
\widetilde{W}_m^{(l)} = \frac{e^{\boldsymbol{\theta_m}^{(l)}\widetilde{\mathcal{T}}^{(l)}_r}}{\sum_{m=1}^{M}e^{\boldsymbol{\theta_m}^{(l)}\widetilde{\mathcal{T}}^{(l)}_r}} \right\}\right.
\end{equation}
where $\boldsymbol{\theta_m}^{(l)}\in \mathcal{R}^{C \times r}$ are the kernel attention parameters and $\sum_{m=1}^MW_m^{(l)} =1$. Based on the weight distribution, we finally obtain the output tensor: 
\begin{equation}
    \mathbf{T}_{final}^{(l)} \overset{\Delta}{=} \sum_{m=1}^M \widetilde{W}_m^{(l)} \otimes \widetilde{T}_m^{(l)} 
    \label{eq:T_new}
\end{equation}
The channel update procedure can be further decomposed as:
\begin{equation}
    \widetilde{W}_m^{(l)} \otimes \widetilde{T}_m^{(l)}  = \left\{
    \widetilde{\omega}_1^{(l)} \otimes\widetilde{\mathcal{T}}_1^{(l)}, \dots, \widetilde{\omega}_{C}^{(l)} \otimes\widetilde{\mathcal{T}}_{C}^{(l)} \right\}
\end{equation}
This shares the same format as the tensor distribution process (equation \ref{eq:weighttensor}) in the temporal attention module but focuses on the channel distribution. The temporal attention parameters $\boldsymbol{\theta_t}^{(l)}$ and kernel attention parameters $\boldsymbol{\theta_r}^{(l)}$, $ \boldsymbol{\theta_m}^{(l)} $ for $l \in [1, L-2]$ are learned through mini-batch stochastic gradient descent (SGD) in the same manner as the TCN unit training \citep{bottou2010large}.

\section{Integration with Dilated Convolutions}\label{dilation}

\begin{figure}[]
  \includegraphics[width=\linewidth]{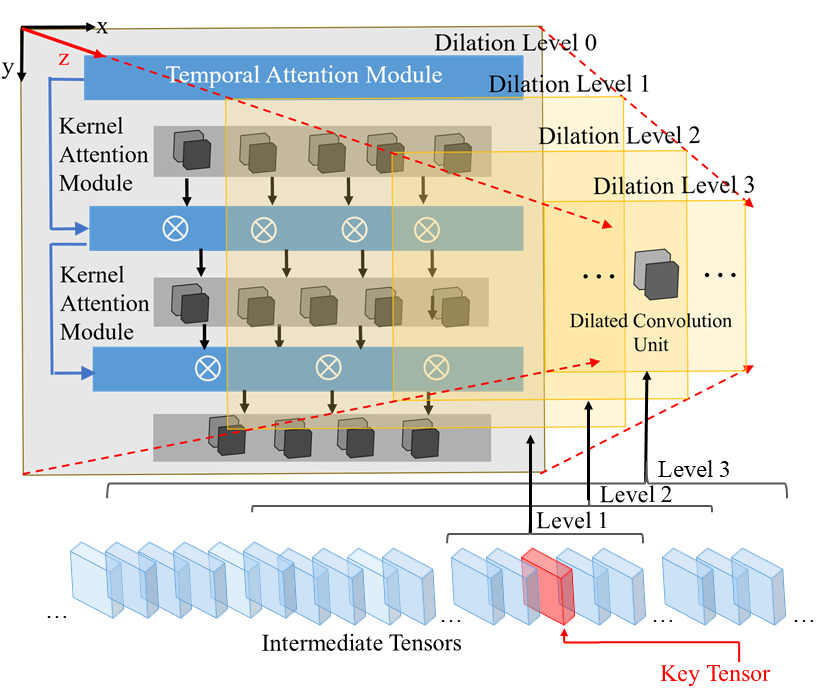}
  \caption{The model of temporal dilated convolution network. As the level index increases, the receptive field over frames (layer index = 0) or tensors (layer index $\geq$  0) increases.  }
\label{fig:3Ddilation}
\end{figure}

For the proposed attention model, a large receptive field is crucial to learn long range temporal relationships across frames, thereby enhancing the estimation consistency. However, with more frames feeding into the network, the number of neural layers increases together with more training parameters. To avoid vanishing gradients or other superfluous layers problems \citep{Martinez2017}, we devise a multi-scale dilation (MDC) strategy by integrating dilated convolutions. 

Fig. \ref{fig:3Ddilation} shows our dilated network architecture. For visualization purpose, we project the network into an $xyz$ space.  The $xy$ plane has the same configuration as the network in
 Fig. \ref{fig:architecture}, with the combination of temporal and kernel attention modules along the $x$ direction, and layers layout along the $y$ direction. As an extension, we place the dilated convolution units (DCUs) along the $z$ direction. This $z$-axis is labeled as \emph{level}s to differ from the \emph{layer} concept along the $y$ direction. As the level index increases, the receptive field grows with increasing dilation size while reducing the  number of DCUs.

\section{Experimental Evaluation}

\begin{figure}
\subfloat[Prototype 1: Layer 4 $\times$ Level 2 (L4$\times$V2)]{\includegraphics[width=\linewidth]{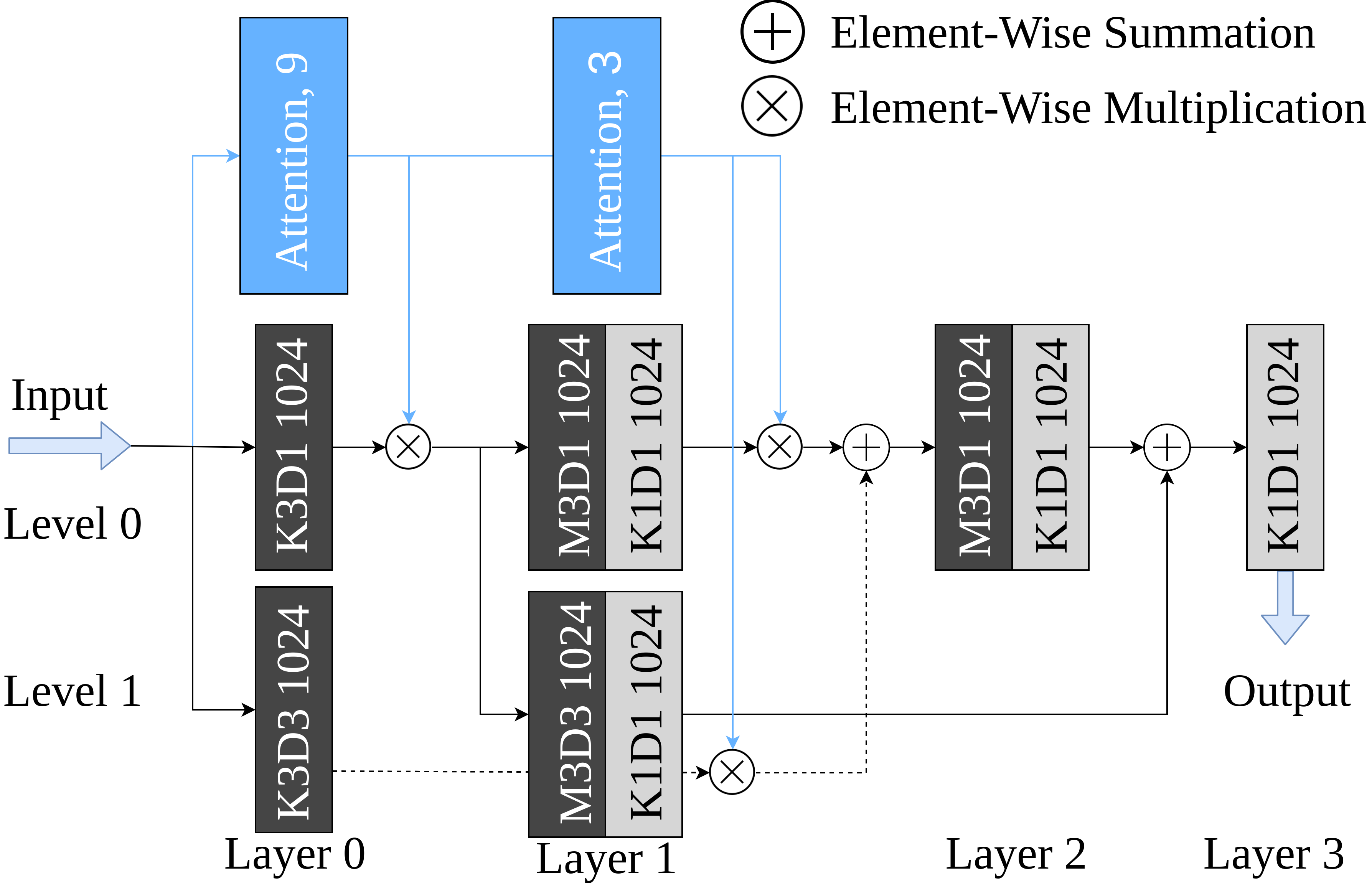}}\\
\subfloat[Prototype 2: Layer 5 $\times$ Level 3 (L5$\times$V3)]{\includegraphics[width=\linewidth]{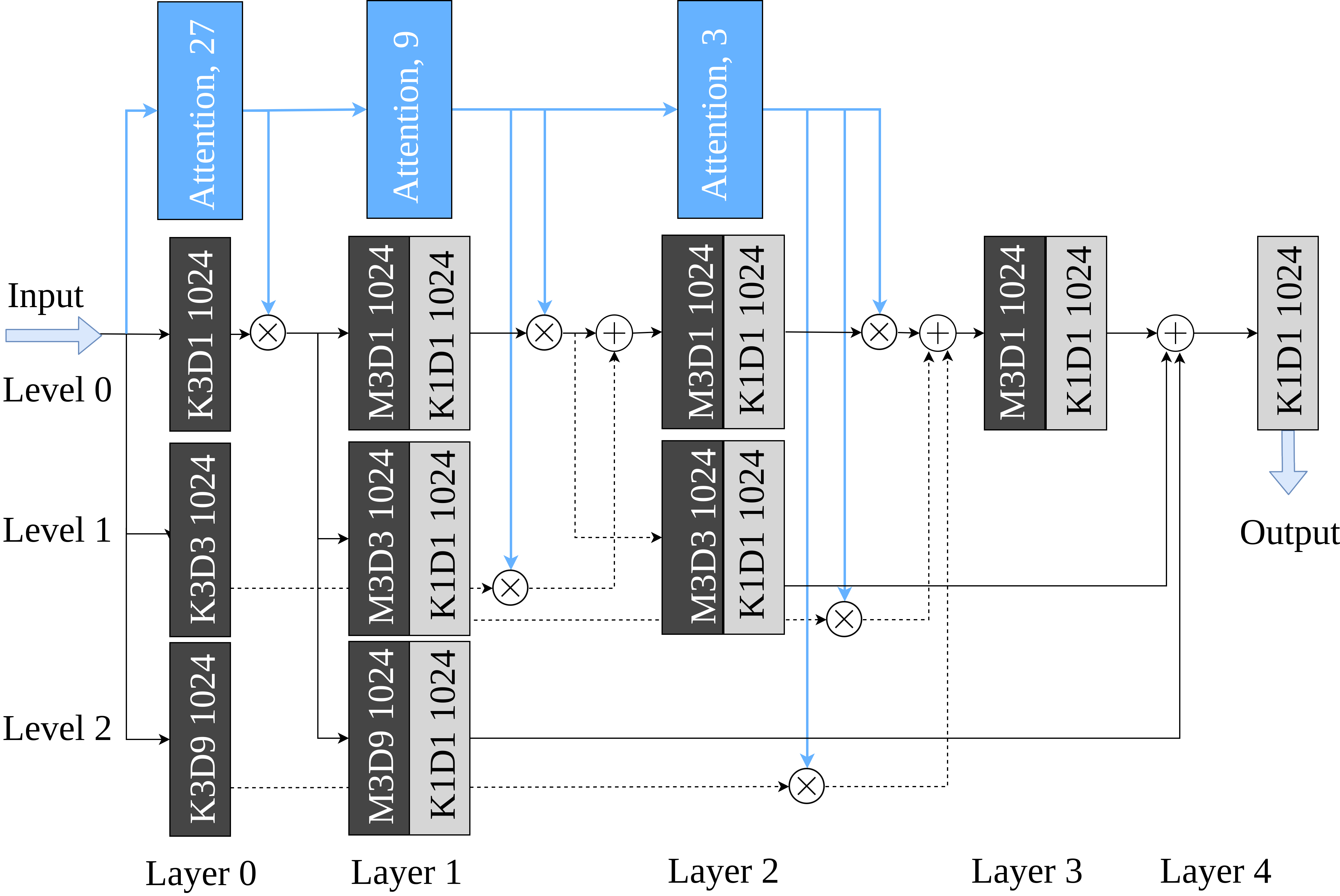}}
\caption{Architectures of input/output data flows across different dilated convolution units. Inside each unit, the numbers represent the unit configuration, e.g. K3D9, 1024 means kernel size is 3, dilation rate is 9, and tensor depth or number of channels is 1024. M3D3, 1024 means TCN units are 3, dilation rate is 9, and tensor depth or number of channels is 1024.}
\label{fig:config}
\end{figure}

This section discusses our system implementation as well as the evaluation results compared to the state-of-the-art techniques by using the standard pose estimation protocols on public datasets.  We first describe the configuration and timings for each functional module, as well as the timings for the run-time algorithm. Ablation studies of the system are conducted by analyzing each component and discuss their performance and limitations. Then we evaluate the estimation accuracy compared to other approaches as well as the ground truth. Finally we  demonstrate the robustness and flexibility of the proposed approach on videos in the wild with various environment complexities and unknown camera settings. Our model is generic and runs on novel users without requiring any offline training or manual preprocessing steps. More extensive evaluation can be found at our lab website \footnote{Demo: \url{https://sites.google.com/a/udayton.edu/jshen1/pose3d}}.

\begin{figure}[ht]
  \includegraphics[width=\linewidth]{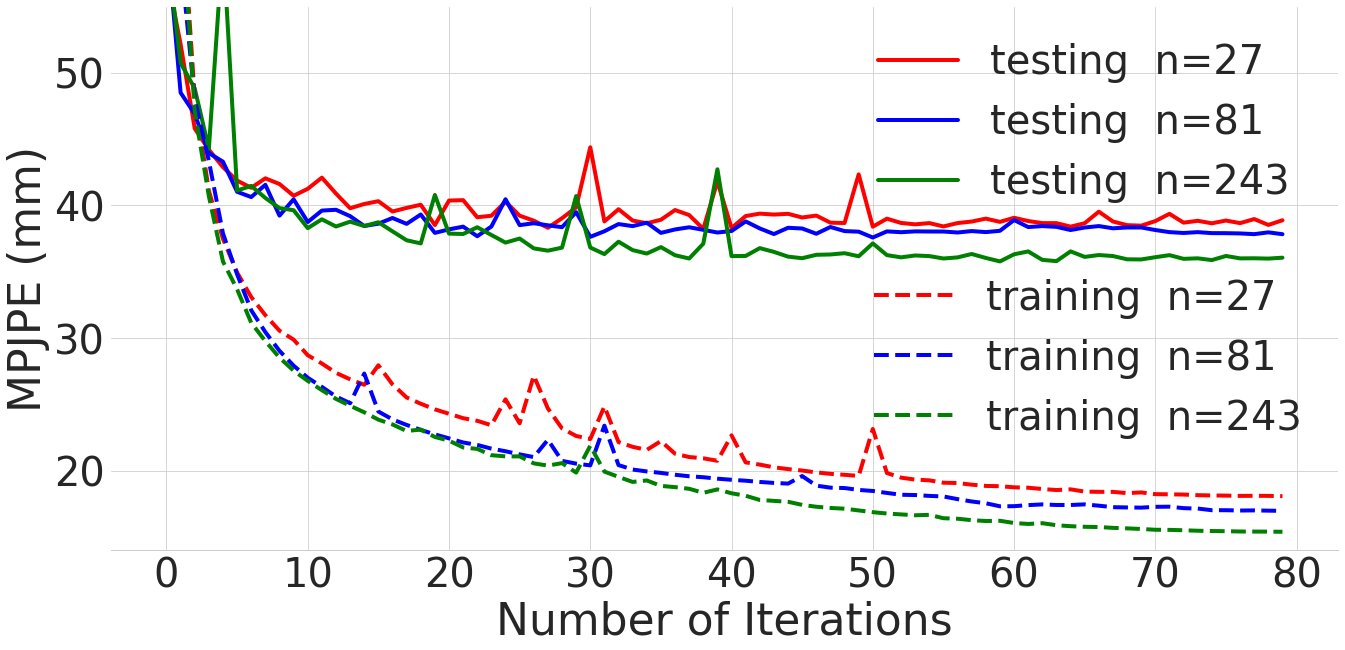}
  \caption{Convergence characteristics for training and testing on three prototypes.}\label{fig:converge}
\end{figure}

\subsection{Configuration and Computational Complexity}
To investigate the practical feasibility of the proposed approach, we implemented three prototypes with different layer $L$ and dilation level $V$ combinations: $L4 \times V2 \times N27$, $L5 \times V3 \times N81$, and $L6\times V4\times N243$, where the last term $N$  indicates the corresponding input frame number. Fig. \ref{fig:config} provides a deeper insight on unit configuration of the prototypes: $L4 \times V2 \times N27$ and $L5 \times V3 \times N81$. By dropping the $x$-axis from Fig. \ref{fig:3Ddilation}, it only displays the level and layer distribute in a 2D view. For simplicity, we use a black/gray rectangle shape to denote the group of TCN units within a layer. At level $0$ , the TCN units are placed by layers along the $y$-axis corresponding to the ones depicted in Fig. \ref{fig:3Ddilation}.  From level $1$,  along the positive $z$-axis, different scaled dilated convolution units are placed. As the level index grows, the number of dilated units decreases due to the increasing receptive fields.

All the prototypes are implemented in native Python (Pytorch 1.0) and tested on a NVIDIA TITAN RTX GPU  without parallel optimization. Despite the difference in layers and levels, all the prototypes present similar convergence rate in training and testing, as shown in Fig. \ref{fig:converge}. With data augmentation, the $L6\times V4$ setting demonstrates the best Mean Per Joint Position Error (MPJPE) performance with approximately $16$ hrs training on $1.6$M frames. The optimizer is Ranger \citep{zhang2019lookahead,liu2019variance}, and the learning rate is 0.001 with decay=0.05 for 80 epoch, the batch size is 1024 and dropout is 0.2. For real-time inference, it can reach $3000$ FPS. 

\begin{table}[ht]
	\begin{center}

			\begin{tabular}{l|c c}
				\toprule
				 Method & Parameters & MPJPE\\
				\midrule
				% Hossain et al.
				ECCV
				\citep{Hossain2018} &16.96M & 41.6\\
				% Pavllo et al.
				CVPR
				\citep{Pavllo2019} (27f) & 8.56M & 40.6\\
				Ours (L4 x V2) & 5.69M & 39.1 \\
				\midrule
				% Pavllo et al.
				CVPR
				\citep{Pavllo2019} (81f) & 12.75M & 38.7\\
				Ours (L5 x V3) & 8.46M & 37.0\\
			    \midrule
			 %   Pavllo et al.
			    CVPR
			    \citep{Pavllo2019} (243f) & 16.95M & 37.8\\
			    Ours (L6 x V4) & 11.22M & 33.4\\
				\bottomrule
			\end{tabular}
	\end{center}
	\caption{Computational complexity performance in terms of the number of involved learning parameters. }
	\label{tb:tb5}
\end{table}

Table \ref{tb:tb5} compares our model with TCN based  semi supervised approach \citep{Pavllo2019}, and the layer normalized LSTM approach \citep{Hossain2018} in terms of the computational complexity. Our model requires fewer parameters for learning the model while achieving better accuracy.  In particular, the input numbers of frames for our three prototypes exactly match the corresponding ones in \citep{Pavllo2019} (i.e., $\#243$, $\#81$, and $\#27$), while ours saves 2M parameters on average. 

\subsection{Datasets and Evaluation Protocols}
Our quantitative evaluation is conducted on two most commonly used datasets: Human3.6M \citep{Ionescu2014} and HumanEva \citep{Sigal2010}. We also applied our approach to some challenging YouTube videos, which include fast motion activities and low-resolution frames. It would be extremely difficult to obtain meaningful 2D detection for those challenging videos collected in the wild. For the Human3.6M, we follow the same training and validation schemes as in the previous works \citep{Martinez2017, Yang2018, Hossain2018, Pavllo2019}. Specifically, subjects S1, S5, S6, S7, and S8 are used for training, and subjects S9 and S11 are used for testing. In the same manner, we conducted training/testing on the HumanEva (a comparatively smaller dataset) with the "Walk" and "Jog" actions performed by subjects S1, S2, and S3.

For both datasets, we use the standard evaluation metrics MPJPE and P-MPJPE to measure the offset between the estimation result and ground-truth (GT) relative to the root node in millimeters \citep{Ionescu2014}. Two protocols are involved in the experiment: {\bf Protocol 1}  computes the mean Euclidean distance across all the joints after aligning the root joints (i.e., \emph{pelvis}) between the predicted and ground-truth poses, referred as MPJPE \citep{fang2018, lee2018, Pavlakos2017, luvizon20182d}.  {\bf Protocol 2} applies additional similarity transformation Procrustes analysis \citep{lepetit2005monocular} to the predicted pose as an enhancement and it is called P-MPJPE \citep{Martinez2017, Hossain2018, Yang2018, Pavllo2019}. In contrast to protocol 1, this evaluation can be more robust to individual joint prediction failure due to the rigid alignment. It is worth mentioning that some researchers also use another protocol by performing a scale alignment on the predicted pose and it is named as N-MPJPE \citep{rhodin2018learning}. Since it has a similar goal as protocol 2 with relatively less transformation, the error usually drops between the outputs produced by protocols 1 \& 2. As such, the accuracy performance should be sufficiently evaluated by using these two protocols.

\subsection{Ablation Studies}\label{ablation_study}
To verify the impact and performance of each component in the network, we conducted ablation experiments on the Human3.6M dataset under $Protocol \#1$. 

\begin{figure}[ht]
\begin{center}
   \includegraphics[width=\linewidth]{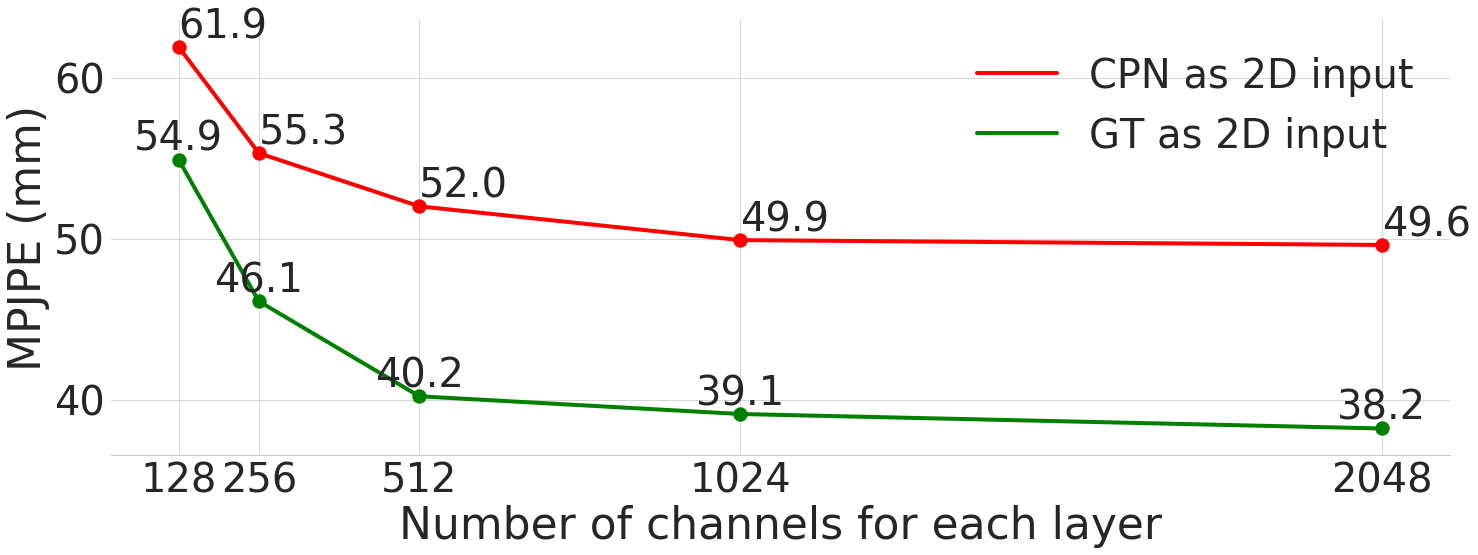}
\end{center}
   \caption{The impact of channel number on MPJPE. CPN: cascaded pyramid network and GT: ground-truth.}\label{channels}
\end{figure}

{\bf TCN Unit Channels: } we first investigated how the channel number $C$ affects the performance between TCN units and temporal attention models. In our test, we used both the CPN and GT as the 2D input. Starting with a receptive field of $n=3\times3\times3=27$, as we increase the channels ($C \leq 512$), the MPJPE drops down significantly. However, the MPJPE changes slowly when $C$ grows between $512$ and $1024$, and remains almost stable afterwards. As shown in Fig. \ref{channels}, with the CPN input, a marginal improvement is yielded from MPJPE $49.9mm$ at $C = 1024$ to $49.6mm$ at $C = 2048$. A similar curve shape can be observed for the GT input. Considering the computation load with more parameters introduced, we chose $C = 1024$ in our experiments.

{\bf Kernel Attention: } Table \ref{kernel_parameters} shows how the setting of different parameters inside the Kernel Attention module impacts the performance under $Protocol \#1$. The left three columns list the main variables. For validation purposes, we divide the configuration into three groups in row-wise. Within each group, we assign different values in one variable while keeping the other two fixed. The items in bold represent the best individual setting for each group. Empirically, we chose the combination of $M = 3$, $G = 8$, and $r = 128$ as the optimal setting (labeled in box). Note,  we select $G = 8$ instead of the individual best assignment $G = 2$, which introduces a larger number of parameters with negligible MPJPE improvement. 

\begin{table}[ht]
    \begin{center}
            \begin{tabular}{c|c|c||c|c}
                \toprule
                Kernels & Groups & Channels  & Parameters & P1  \\
                \midrule
                    M=1     & G=1            & -          & 16.95M & 37.8 \\
                    M=2     & G=8            & r=128        & 9.14M  & 37.1 \\
                    \boxed{\textbf{M=3}}     & G=8           & r=128       & 11.22M & \boxed{\textbf{35.5}} \\
                    M=4     & G=8            & r=128        & 13.36M & 38.0\\
            \midrule
                    M=3     & G=1            & r=128        & 44.25M &  37.4  \\
                    M=3     & \textbf{G=2}            & r=128        & 25.41M & \textbf{35.3} \\
                    M=3     & G=4            & r=128        & 15.97M & 35.6 \\
                    M=3     & \boxed{G=8}           & r=128       & 11.25M & \boxed{35.5} \\
                    M=3     & G=16           & r=128        & 8.89M  & 37.3 \\
            \midrule
                    M=3     & G=8            & r=64         & 10.20M & 35.9 \\
                    M=3     & G=8           & \boxed{\textbf{r=128}}       & 11.25M & \boxed{\textbf{35.5}} \\
                    M=3     & G=8            & r=256        & 13.35M & 36.2 \\
                \bottomrule
            \end{tabular}
    \end{center}
    \caption{Ablation study on different parameters in our kernel attention model. Here, we use receptive field $n=3\times3\times3\times3\times3=243$. The evaluation is performed on Human3.6M under $Protocol \#1$ with MPJPE (mm).}
    \label{kernel_parameters}
\end{table}

In Table \ref{kernel_receptive}, we discuss the choice of different types of receptive fields and how it affects the network performance. The first column shows various layer configurations, which generate different receptive fields, ranging from $n = 27$ to $n = 1029$. To validate the impact of $n$, we fix the other parameters, i.e. $M = 3$, $G = 8$, $r = 128$. Note that for a network with smaller number of layers (e.g. $L = 3$), a larger receptive field may reduce the error more effectively. For example, increasing the receptive field from $n = 3\times3\times3 = 27$ to $n = 3\times3\times7 = 147$, the MPJPE drops from $40.6$ to $36.8$ . However, for a deeper network, a larger receptive field may not be always optimal, e.g. when $n = 1029$, MPJPE $= 37.0$. Empirically, we obtained the best performance with the setting of $n = 243$ and $L = 5$, as indicated in the last row.

\begin{table}[ht]
    \begin{center}
    \scalebox{0.8}{
            \begin{tabular}{c|c|c|c||c|c}
                \toprule
                Receptive fields& Kernels & Groups & Channels  & Parameters & P1  \\
                \midrule
                    3$\times$3$\times$3=27  & M=1   & G=1            & -          & 8.56M & 40.6 \\
                    \midrule
                    3$\times$3$\times$3=27  & M=2   & G=4            & r=128      & 6.21M & 40.0 \\
                    3$\times$5$\times$3=45  & M=2   & G=4        & r=128      & 6.21M & 39.9 \\
                    3$\times$5$\times$5=75  & M=2   & G=4        & r=128      & 6.21M & 38.5 \\
                    \midrule
                    3$\times$3$\times$3=27  & M=3   &G=8           & r=128      & 5.69M & 39.5 \\
                    3$\times$5$\times$3=45  & M=3   & G=8           & r=128      & 5.69M & 39.2 \\
                    3$\times$5$\times$5=75  & M=3   & G=8          & r=128      & 5.69M & 38.2 \\
                    3$\times$7$\times$7=147 & M=3   & G=8           & r=128      & 5.69M & 36.8 \\
                    \midrule
                    3$\times$3$\times$3$\times$3=81&  M=3  &  G=8           &  r=128      &  8.46M &  37.8 \\
                    3$\times$5$\times$5$\times$5=375 & M=3 & G=8           & r =128    & 8.46M  & 36.6 \\
                    3$\times$7$\times$7$\times$7=1029 & M=3   & G=8         & r=128    & 8.46M & 37.0 \\
                    \midrule
                    \textbf{3}$\times$\textbf{3}$\times$\textbf{3}$\times$\textbf{3}$\times$\textbf{3}=\textbf{243}&  M=3  &  G=8            &  r=128      &  11.25M &  \textbf{35.5} \\
                \bottomrule
            \end{tabular}
            }
    \end{center}
    \caption{Ablation study on different receptive fields in our kernel attention model. The evaluation is performed on Human3.6M under $Protocol \#1$ with MPJPE (mm).}
    \label{kernel_receptive}
\end{table}

{\bf Multi-Scale Dilation: } To evaluate the impact of the dilation component on the network,  we tested the system with and without dilation and compared their individual outcomes. In the same way, the GT and CPN  2D detectors are used as input and being tested on the Human3.6M dataset under $Protocol \#1$. Table \ref{tb:model} demonstrates the integration of attention, and multi-scale dilation components surpass their individual performance with the minimum MPJPE for all the three prototypes. We also found the attention model makes an increasingly significant contribution as the layer number grows. This is because more layers lead to a larger receptive field, allowing the multi-scale dilation to capture long-term dependency across frames. The effect is more noticeable when fast motion or self-occlusion present in videos. 

\begin{table}[ht]
    \begin{center}
    \scalebox{0.9}{
            \begin{tabular}{l|c|c|c}
                \toprule
                \diagbox[width=15em]{Method}{Model} & $n=27$ & $n=81$ & $n=243$ \\
                \midrule
                Attention model (CPN) & 49.1 & 47.2 & 45.7\\
                Multi-Scale Dilation model (CPN) & 50.3 & 49.8 & 49.1\\
                Attention and Dilation (CPN) & 49.0 & 46.5 & 45.1\\
                \midrule
                Attention model (GT) & 39.5 & 37.8 & 35.5\\                
                Multi-Scale Dilation model (GT) & 39.2 & 37.2 & 35.3\\
                Attention and Dilation (GT) & 38.9 & 36.2 & 33.4\\
                \bottomrule
            \end{tabular}
            }
    \end{center}
    \caption{Ablation study on different components in our method. The evaluation is performed on Human3.6M under $Protocol \#1$ with MPJPE (mm).}
    \label{tb:model}
\end{table}

{\bf Step by step performance enhancement: } Here we list all the steps and additional modules used to obtain the performance. The step-by-step gains brought by each component are illustrated in Table \ref{tb:incements}.
\begin{table}[ht]
    \begin{center}
            \begin{tabular}{l|c|c}
                \toprule
                \diagbox[width=15em]{Model}{Method} & CPN & GT \\
                \midrule
                Baseline 27 frames & 51.2 & 40.6\\
                \midrule
                + Receptive field(243 frames)& 49.2 & 37.8\\
                                \midrule
                + Attention & 47.9 & 35.5\\
                                \midrule
                + Dilation & 47.2 & 34.7\\
                                \midrule
                + Project on 2D & 47.0 & 34.5\\
                                \midrule
                + 2D pose enhance & 46.5 & -\\
                                \midrule
                + Data augment & 44.8 & 33.4\\
                \bottomrule
            \end{tabular}
    \end{center}
    \caption{Ablation study on different components in our method. The evaluation is performed on Human3.6M under $Protocol \#1$ with MPJPE (mm).}
    \label{tb:incements}
\end{table}

\subsection{Comparison with State-of-the-Art}
\begin{table*}
\begin{center}
	\scalebox{0.85}{
		\begin{tabular}{l|ccccccccccccccc|c}
			\toprule
			Method & Dir. & Disc. & Eat & Greet & Phone & Photo & Pose & Pur. & Sit & SitD. & Smoke & Wait & WalkD. & Walk & WalkT.& Avg \\
			\midrule
% 			Martinez et al. ICCV'17
            % ICCV
            \citep{Martinez2017} & 51.8 & 56.2 & 58.1 & 59.0 & 69.5 & 78.4 & 55.2 & 58.1 & 74.0 & 94.6 & 62.3 & 59.1 & 65.1 & 49.5 & 52.4 & 62.9 \\
% 			Fang et al. AAAI'18
            % AAAI
            \citep{fang2018} & 50.1 & 54.3 & 57.0 & 57.1 & 66.6 & 73.3 & 53.4 & 55.7 & 72.8 & 88.6 & 60.3 & 57.7 & 62.7 & 47.5 & 50.6 & 60.4\\
% 			Yang et al. CVPR'18
            % CVPR
			\citep{Yang2018} & 51.5 & 58.9 & 50.4 & 57.0 & 62.1 & 65.4 & 49.8 & 52.7 & 69.2 & 85.2 & 57.4 & 58.4 & \underline{43.6} & 60.1 & 47.7 & 58.6\\
% 			Pavlakos et al. CVPR'18
            % CVPR
			\citep{Pavlakos2017} & 48.5 & 54.4 & 54.4 & 52.0 & 59.4 & 65.3 & 49.9 & 52.9 & 65.8 & 71.1 & 56.6 & 52.9 & 60.9 & 44.7 & 47.8 & 56.2\\
% 			Luvizon et al. CVPR’18
% 			CVPR
			\citep{luvizon2018} & 49.2 & 51.6 & 47.6 & 50.5 & 51.8 & 60.3 & 48.5 & 51.7 & 61.5 & 70.9 & 53.7 & 48.9 & 57.9 & 44.4 & 48.9 & 53.2\\
% 			Hossain et al. ECCV'18
% 			ECCV
			\citep{Hossain2018} & 48.4 & 50.7 & 57.2 & 55.2 & 63.1 & 72.6 & 53.0 & 51.7 & 66.1 & 80.9 & 59.0 & 57.3 & 62.4 & 46.6 & 49.6 & 58.3\\
% 			Lee et al. ECCV'18
% 			ECCV
			\citep{lee2018} & \textbf{40.2 }& 49.2 & 47.8 & 52.6 & 50.1 & 75.0 & 50.2 & 43.0 &  \underline{55.8} & 73.9 & 54.1 & 55.6 & 58.2 & 43.3 & 43.3 & 52.8\\
% 			Dabral et al. ECCV'18
% 			ECCV
			\citep{dabral2018learning} & 44.8 & 50.4 & 44.7 & 49.0 & 52.9 & 61.4 & 43.5 & 45.5 & 63.1 & 87.3 & 51.7 & 48.5 & 52.2 & 37.6 & 41.9  & 52.1\\
% 			Zhao et al. CVPR'19
% 			CVPR
			\citep{zhao2019semantic} & 47.3 & 60.7 & 51.4 & 60.5 & 61.1 & 49.9 & 47.3 & 68.1 & 86.2 & 55.0 & 67.8 & 61.0 & \textbf{42.1} & 60.6 & 45.3 & 57.6\\
% 			Pavllo et al. CVPR'19
% 			CVPR
			\citep{Pavllo2019} & 45.2 & \underline{46.7} & \underline{43.3} & \underline{45.6} & \underline{48.1} & \underline{55.1} & \underline{44.6} & \underline{44.3} & 57.3 & \underline{65.8} & \underline{47.1} & \underline{44.0} & 49.0 & \underline{32.8} & \underline{33.9} & \underline{46.8}\\
			\citep{cheng2019occlusion} & - & - & - & - & - & - & - & - & - & - & - & - & - & - & - & $\mathbf{44.8}^{+}$\\
			\midrule
			Ours (n=243 CPN) &  \underline{41.8} &\textbf{44.0} & \textbf{41.1} & \textbf{43.9} & \textbf{47.4} &\textbf{54.1} & \textbf{42.1} &  \textbf{42.2} & \textbf{55.3} & \textbf{63.6} & \textbf{45.3} & \textbf{42.7} & 45.3 & \textbf{31.3} & \textbf{32.2} & \textbf{44.8}\\
          \midrule
			\midrule
% 			Martinez et al. ICCV'17
% 			ICCV
			\citep{Martinez2017} & 37.7 & 44.4 & 40.3 & 42.1 & 48.2 & 54.9 & 44.4 & 42.1 & 54.6 & 58.0 & 45.1 & 46.4 & 47.6 & 36.4 & 40.4 & 45.5 \\
% 			Hossain et al. ECCV'18
% 			ECCV
			\citep{Hossain2018} & 35.2 & 40.8 & 37.2 & 37.4 & 43.2 & 44.0 & \underline{38.9} & \underline{35.6} & \underline{42.3} & 44.6 & 39.7 & 39.7 & 40.2 & 32.8 & 35.5 & 39.2\\
% 			Lee et al. ECCV'18
% 			ECCV
			\citep{lee2018} & \textbf{32.1} & \underline{36.6} &  34.4& 37.8 & 44.5 & 49.9 & 40.9 & 36.2 & 44.1 & 45.6 & \underline{35.3} & \underline{35.9} & \underline{37.6} & 30.3 & 35.5 & 38.4\\
% 			Zhao et al. CVPR'19
            % CVPR
            \citep{zhao2019semantic} & 37.8 & 49.4 & 37.6 & 40.9 & 45.1 & \underline{41.4} & 40.1 & 48.3 & 50.1 & \underline{42.2} & 53.5 & 44.3 & 40.5 & 47.3 & 39.0 & 43.8\\
% 			Pavllo et al. CVPR'19
% 			CVPR
			\citep{Pavllo2019} & 35.2 & 40.2 & \underline{32.7} & \underline{35.7} & \underline{38.2} & 45.5 & 40.6 & 36.1 & 48.8 & 47.3 & 37.8 & 39.7 & 38.7 & \underline{ 27.8} & \underline{29.5} & \underline{37.8}\\
			\midrule
			Ours (n=243 GT) & \underline{33.0} &  \textbf{35.7} & \textbf{31.7} & \textbf{32.4} & \textbf{32.1} & \textbf{36.5} & \textbf{37.2} &\textbf{32.6} &\textbf{40.7} & \textbf{41.4} & \textbf{32.6} & \textbf{33.1} & \textbf{30.9} & \textbf{24.9} & \textbf{25.9} & \textbf{33.4}\\

			\bottomrule
		\end{tabular}
		}
\end{center}
\caption{Protocol 1: Reconstruction Error on Human3.6M.  Top-half: input 2D joints are acquired by detection; bottom-half: input 2D joints with ground-truth. {\bf (CPN)} - cascaded pyramid network, {\bf (GT)} - ground-truth, ($+$) - the result without data augmentation using virtual cameras}
\label{tb:tb1}
\end{table*}

\begin{table*}
	\begin{center}
		\scalebox{0.85}{
			\begin{tabular}{l|ccccccccccccccc|c}
				\toprule
				Method & Dir. & Disc. & Eat & Greet & Phone & Photo & Pose & Pur. & Sit & SitD. & Smoke & Wait & WalkD. & Walk & WalkT.& Avg \\
				\midrule
				% Martinez et al. ICCV'17
				% ICCV
				\citep{Martinez2017} & 39.5 & 43.2 & 46.4 & 47.0 & 51.0 & 56.0 & 41.4 & 40.6 & 56.5 & 69.4 & 49.2 & 45.0 & 49.5 & 38.0 & 43.1 & 47.7\\
				
				% Fang et al. AAAI'18
				% AAAI
				\citep{fang2018} & 38.2 & 41.7 & 43.7 & 44.9 & 48.5 & 55.3 & 40.2 & 38.2 & 54.5 & 64.4 & 47.2 & 44.3 & 47.3 & 36.7 & 41.7 & 45.7\\
				
				% Hossain et al. ECCV'18
				% ECCV
				\citep{Hossain2018} & 35.7 & 39.3 & 44.6 & 43.0 & 47.2 & 54.0 & 38.3 & 37.5 & 51.6 & 61.3 & 46.5 &  41.4 & 47.3 & 34.2 & 39.4 & 44.1\\
				
				% Pavlakos et al. CVPR'18
				% CVPR
				\citep{Pavlakos2017} & 34.7 & 39.8 & 41.8 & 38.6 & 42.5 & 47.5 & 38.0 & 36.6 & 50.7 & 56.8 & 42.6 & 39.6 & 43.9 & 32.1 & 36.5 & 41.8\\
				
				% Yang et al. CVPR'18
				% CVPR
				\citep{Yang2018} & \textbf{26.9} & \underline{30.9} & 36.3 & 39.9 & 43.9 & 47.4 & \textbf{28.8} & \textbf{29.4} & \textbf{36.9} & 58.4 & 41.5 & \textbf{30.5} &\underline{29.5} & 42.5 & 32.2 & 37.7\\
                
				% Dabral et al. ECCV'18
				% ECCV
				\citep{dabral2018learning} & 28.0 & \textbf{30.7} & 39.1 & \textbf{34.4} & 37.1 & \textbf{28.9} & \underline{31.2} & 39.3 & 60.6 & \textbf{39.3} & 44.8 & \underline{31.1} &\textbf{ 25.3} & 37.8 & 28.4 & \underline{36.3}\\
                
				% CVPR
				\citep{Pavllo2019} & 34.1 & 36.1 & \underline{34.4} & 37.2 & \underline{36.4} & 42.2 & 34.4 & 33.6 & 45.0 & 52.5 & \underline{37.4} & 33.8 & 37.8 & \underline{25.6} & \underline{27.3} & 36.5\\
				
				\midrule
				Ours (n=243 CPN) & \underline{32.3} & 35.2& \textbf{33.3} & \underline{35.8}&\textbf{ 35.9} &\underline{41.5}& 33.2 & \underline{32.7} &\underline{44.6} & \underline{50.9} & \textbf{37.0} &32.4 & 37.0& \textbf{25.2} & \textbf{27.2} & \textbf{35.6}\\
				\bottomrule
			\end{tabular}
		}
		
	\end{center}
	\caption{Protocol 2: Reconstruction Error on Human3.6M with similarity transformation. }
	\label{tb:tb2}
\end{table*}

\begin{table}
	\begin{center}
		\scalebox{0.8}{
			\begin{tabular}{l|ccc|ccc|c}
				\toprule
				 & & Walk & & & Jog & & \\
				 & S1 & S2 & S3 & S1 & S2 & S3 & Avg\\
				\midrule
				% Pavlakos et al.
				\citep{Pavlakos2017} (MA) & 22.3 & 19.5 & 29.7 & 28.9 & 21.9 & 23.8 & 24.35 \\
				% Martinez et al. 
				\citep{Martinez2017} (SA) & 19.7 & 17.4 & 46.8 & 26.9 & 18.2 & 18.6 & 24.6 \\
				% Lee et al. 
				\citep{lee2018} (MA) & 18.6 & 19.9 & 30.5 &25.7 & 16.8 & 17.7 & 21.5\\
				% Pavllo et al.
				\citep{Pavllo2019}(MA) & \underline{13.4} & \underline{10.2} & \underline{27.2} & \underline{17.1} & \underline{13.1} & \underline{13.8} & \underline{15.8}\\
				\midrule
				Ours (n=27 MA) & \textbf{13.1} & \textbf{9.8} & \textbf{26.8} & \textbf{16.9} & \textbf{12.8} & \textbf{13.3} & \textbf{15.4}\\
				\bottomrule
			\end{tabular}
		}
	\end{center}
	\caption{Protocol 2: Reconstruction Error on HumanEva. {\bf (MA)} - multi-action model,  {\bf (SA)} - single action model }
	\label{tb:tb3}
\end{table}

\begin{figure}
    \centering
    \subfloat[Protocol 1: joint error analysis across frames in Human3.6M \emph{Walking S9} left ankle.]{\includegraphics[width= \linewidth]{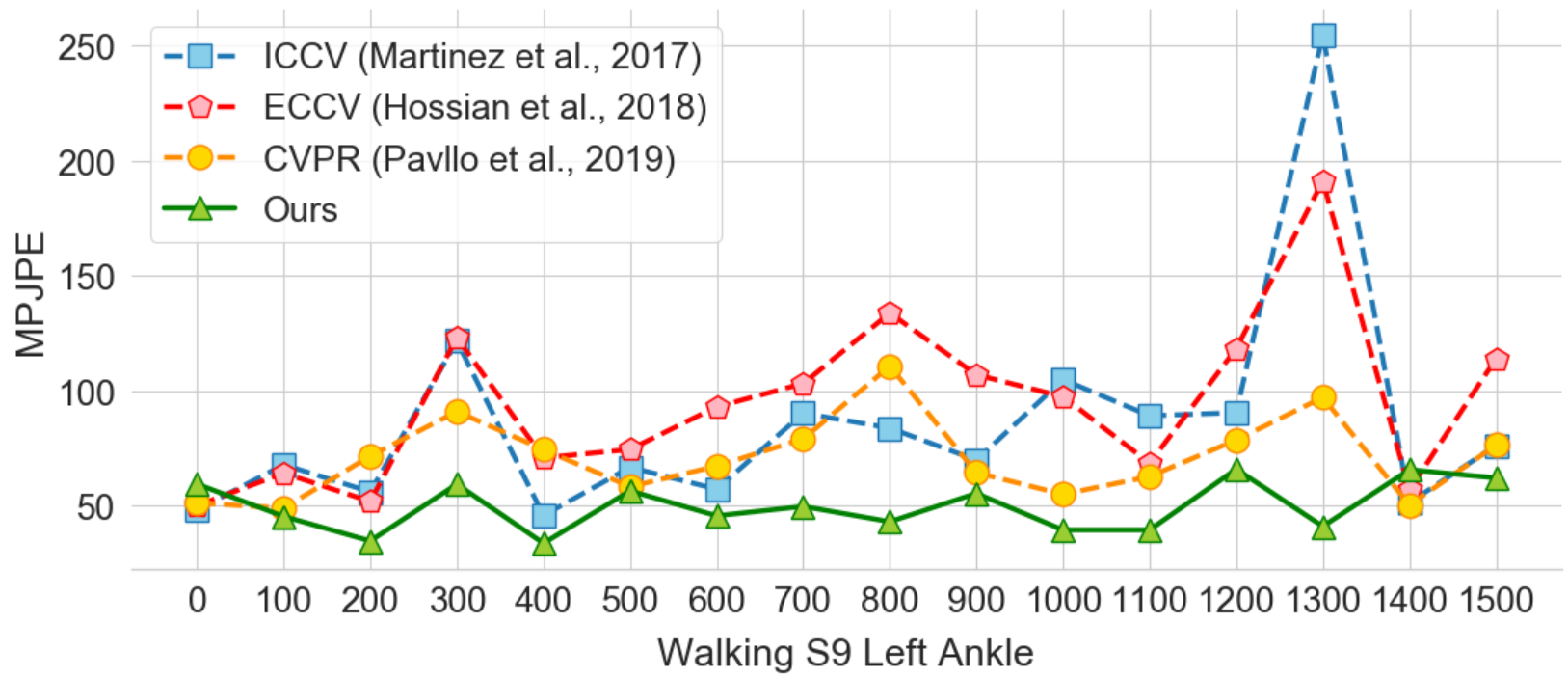}\label{fig:joint1}}\\
    \subfloat[Protocol 1: joint error analysis across frames in Human3.6M \emph{Smoking S9} left elbow. ]{\includegraphics[width= \linewidth]{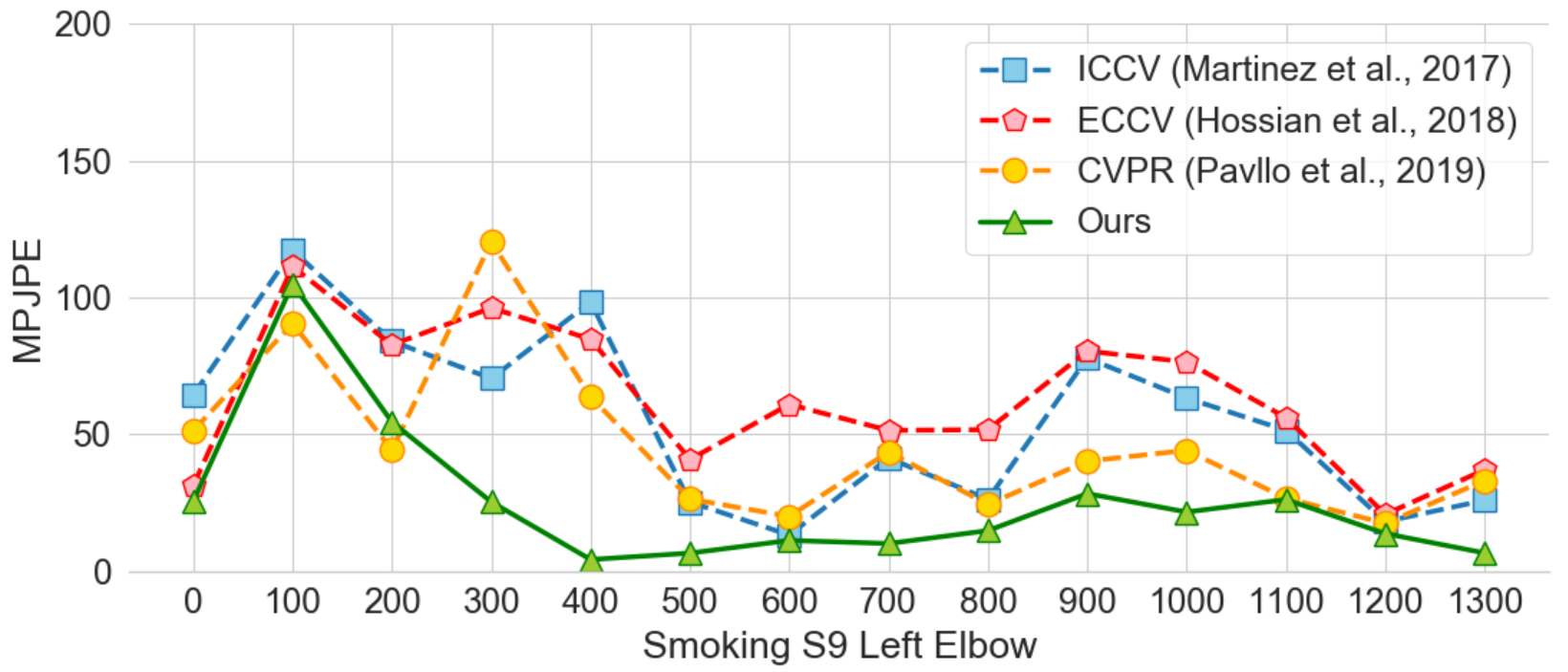}\label{fig:joint2}}
    \caption{Joint-wise analysis across frames.}\label{fig:joint-wise-across-frames}
\end{figure}

\begin{figure*}
    \centering
    \includegraphics[width=\linewidth]{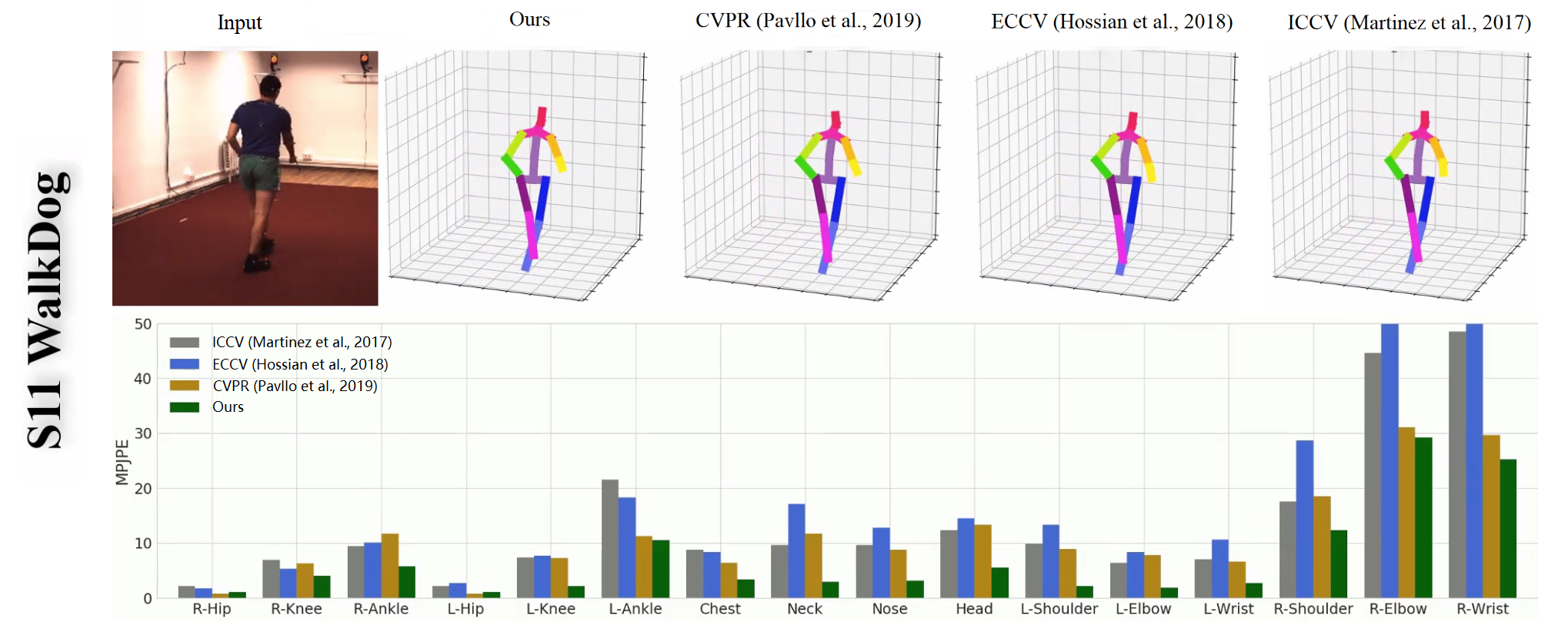}
    \caption{Individual joint MPJPE comparison with state-of-the-art.}
    \label{fig:joint-wise-one-frame}
\end{figure*}

In this subsection, we systematically analyze the performance of our proposed method by comparing it with state-of-the-art. To fairly evaluate the accuracy, we use the same training and testing datasets as others.  Tables \ref{tb:tb1} - \ref{tb:tb3} demonstrate the comparison by following Protocol 1 and 2. Tables \ref{tb:tb1} and \ref{tb:tb2} illustrate the Human3.6M results and Table \ref{tb:tb3} illustrates the results of HumanEva. The results of each method are displayed in row-wise. Each column indicates a different pose scene, e.g., walking, eating, etc. We highlight the best and second best results in each column in bold and underline formats respectively. The last column of each table shows the average results across all the different pose scenes. Note that our model outperforms all the existing approaches by reaching a minimum average error of $48.6$mm in MPJPE and $37.7$mm in P-MPJPE.  
Admittedly, for some pose scenes, e.g., Phone, Eat, our method does not achieve the best performance. This could be due to the nature of the particular activities, for example, if the less noticeable motion or only upper-body movement are involved, limited information is fed into the attention layers to learn tensor distributions. However, if one considers all the scenarios, our overall performance demonstrates higher accuracy than other methods by a fair margin. In particular, under protocol 1, our model reduces the best-reported error rate of MPJPE  \citep{Pavllo2019} by approximate 3\% using ground truth 2D pose as the input.

To further demonstrate the efficacy, we evaluated the performance and advantage of our approach in three aspects:
\begin{enumerate}
\item  \textbf{Joint-wise}: Analyzing the accuracy of individual joint measurement with MPJPE comparison 
\item  \textbf{Frame-wise}: Tracking the average MPJPE of all the joints across frames
\item  \textbf{Re-targeting-wise}: Applying motion-retargeting by transferring the estimated pose to a 3D avatar
\end{enumerate}
We compare our approach with three state-of-the-art techniques, which represent the best reported results on monocular video-based \emph{2D-to-3D estimation} to date:  the deep feedforward 2D-to-3D network \citep{Martinez2017}, the layer-normalized LSTM based algorithm \citep{Hossain2018}, and the dilated temporal convolution with semi-supervised training \citep{Pavllo2019}. Fig. \ref{fig:joint-wise-one-frame} demonstrates the joint-wise MPJPE for a selected frame from the \emph{WalkDog S11} data. The top row shows the input 2D color image and its corresponding estimated 3D poses by other methods. The histograms in the second row show the quantified measurement on each joint, e.g., R-Knee, Nose, Neck. Note that our approach, indicated by the green bar, achieves minimum MPJPE among all the other methods in most of the joints. To further validate the accuracy, we trace these individual joints across frames in the corresponding video sequence and measure their MPJPE in the temporal space. Fig. \ref{fig:joint-wise-across-frames} plots the MPJPE curves over time (around 1400 frames) on two selected joints: the left ankle from \emph{Walking S9} and the left elbow from \emph{Smoking S9}. Compared to the recent works \citep{Martinez2017, Hossain2018, Pavllo2019}, our results yield low errors consistently through learning the long-range dependencies using the multi-scale dilation convolution. 

\begin{figure}[ht]
 \subfloat[Pavllo et al., 2019]{\includegraphics[width= 1.1in]{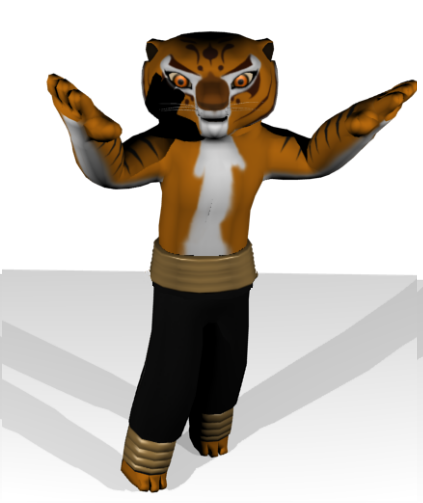}}
 \subfloat[Ground truth]{\includegraphics[width= 1.1in]{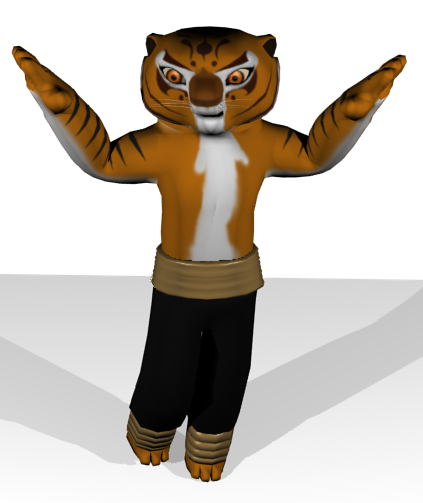}}
 \subfloat[Ours]{\includegraphics[width=1.1in]{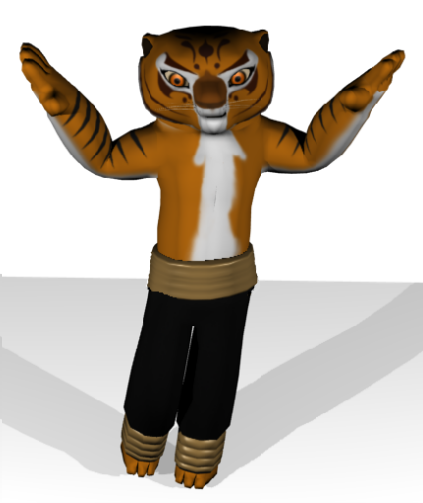}}\label{fig:compare1c}
 \subfloat{\includegraphics[width= 3.3in]{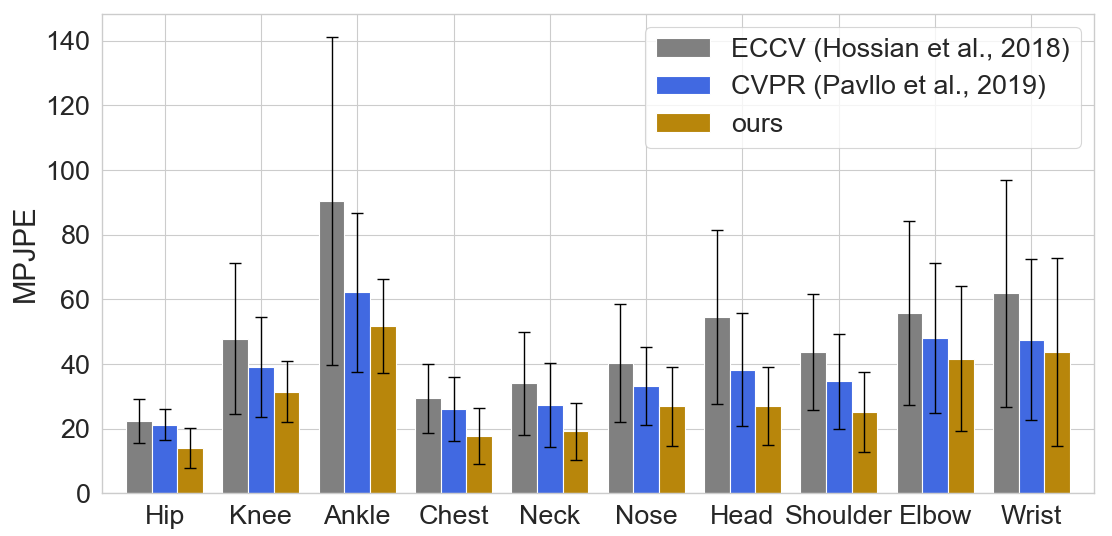}\\}
 \caption{Comparison results: (top): side-by-side views of motion retargeting results on a 3D avatar; the source is from frame 857 of \emph{walking S9} and frame 475 \emph{posing S9} in Human3.6M. (bottom): the average joint error comparison across all the frames of the video walking S9 \citep{Pavllo2019}. }
 \label{fig:compare1}
 \end{figure}

\begin{figure*}
    \centering
    \includegraphics[width=\linewidth]{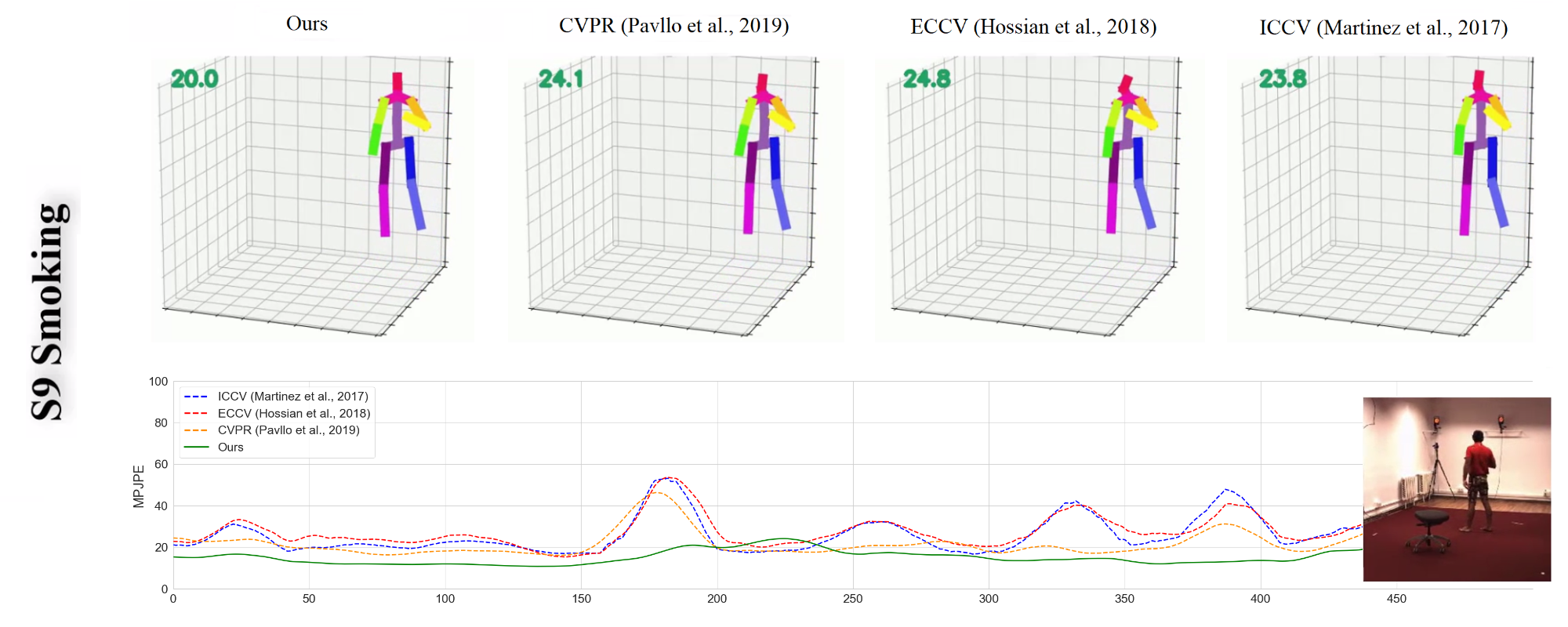}\\
    \includegraphics[width=\linewidth]{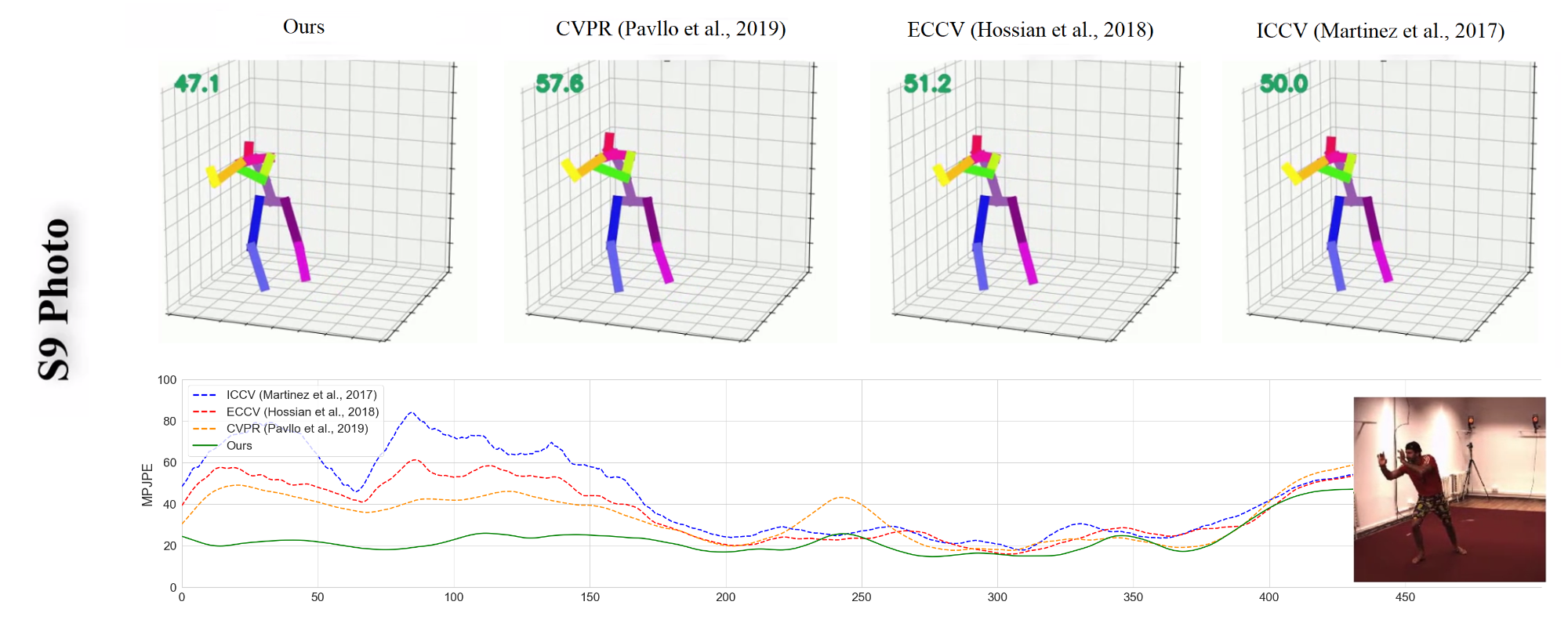}
    \caption{Frame-wise comparasion with state-of-the-art results.}
    \label{fig:frame-wise-whole-video}
\end{figure*}

In light of possible biases and uncertainties that individual joint may introduce, we perform frame-wise analysis by taking the average MPJPE of all the joint estimation in each frame and measure how it changes through a video sequence. Fig. \ref{fig:frame-wise-whole-video} shows the testing results on two scenes of the Human3D dataset: \emph{smoking S9} and \emph{photo S9}. For each scene, the top row presents the estimated 3D pose results from the same frame produced by different methods. Though it is hard to see the difference from the single frame, from the MPJPE (the green number on the top-left corner of each pose result), our attention-based model delivers the best result. In the second row of each scene, We  trace these average joint errors across all of the frames in the corresponding video sequence. Our results maintain low MPJPE compared to the other methods. 
 
\begin{figure*}
    \centering
    \includegraphics[width= \linewidth]{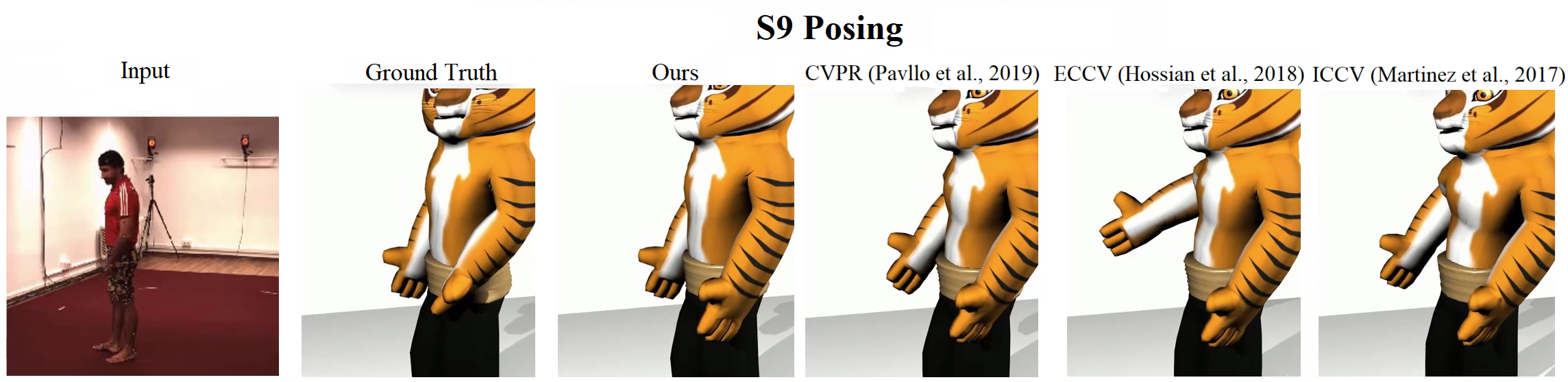}
    \includegraphics[width= \linewidth]{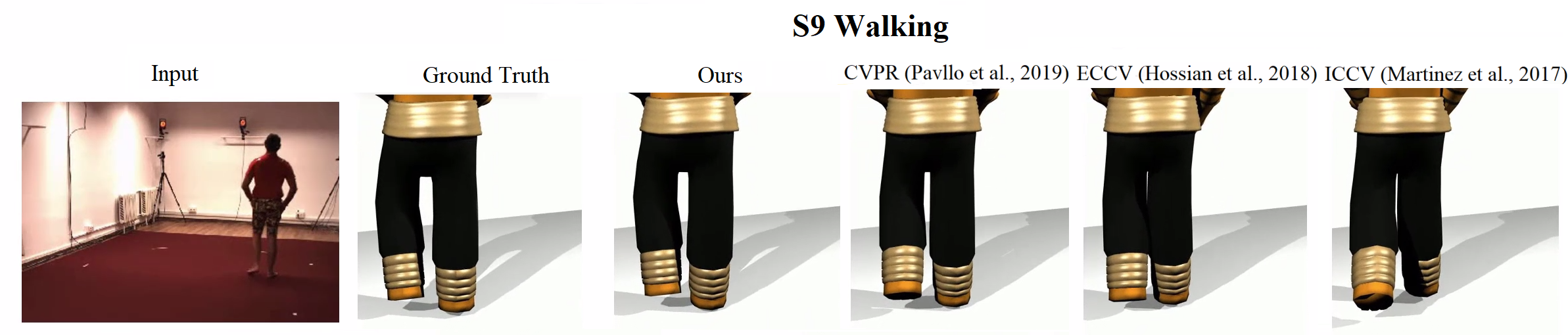}
    \caption{Comparison with state-of-the-art results on motion re-targeting model.}
    \label{fig:retarget-frame-wise}
\end{figure*}

To visually demonstrate the significance of the estimation improvement, we apply animation retargeting to a 3D avatar by synthesizing the captured motion from the same frame of the \emph{Walking S9} and  \emph{Posing S9} sequences as shown in Fig. \ref{fig:compare1}. With the support of additional mesh surface driven by the pose, it helps magnify the degree of body part arrangement that enhance the contrast of estimation. From the side-by-side comparisons, one can easily see the difference between the rendered results against the ground truth. Specifically, the shadows of the legs and the right hand are differently rendered due to the erroneous pose estimated using the method in \citep{Pavllo2019} while ours stay more aligned to the ground truth. The quantified MPJPE for each joint estimation is shown in the correponding histograms right below it. Fig. \ref{fig:retarget-frame-wise} shows more retargeting results on the same dataset for different frames. The zoom-in views illustrate the details of the animated characters of different pose configurations. For the \emph{Posing S9} in the first row, our results bear the closest similarity as the ground truth with the right arm of the character naturally hanging down the side of the body, while others present more distinct arm gesture. The second row of Fig. \ref{fig:retarget-frame-wise} demonstrates the improvement of our approach on leg movement prediction with optimistic estimate on the two legs relative positions. Note that this is just one selected frame from the walking sequence, which is a common body activity involving the alternate of left and right legs in a repetitive manner. Accurate and consistent part detection is crucial to deliver  smooth motion sequences without any jittering effect in 3D pose reconstruction. 

\subsubsection*{\bf 2D Detection}

\begin{table}
	\begin{center}
			\begin{tabular}{l|c c}
				\toprule
				Method & P1 & P2 \\
				\midrule
				
				% Martinez et al.
				ICCV
				\citep{Martinez2017}(SH PT) & 67.5 & 52.5\\
				ours (L6 x V4 SH PT) & \textbf{57.3} & \textbf{45.5} \\
				
               \midrule
				% Martinez et al.
				ICCV
				\citep{Martinez2017}(SH FT) & 62.9 & 47.7 \\
				% Hossain et al.
				ECCV
				\citep{Hossain2018}(SH FT) & 58.3 & 44.2 \\
				% Zhao et al.
				CVPR
				\citep{Zhao2019}(SH FT) & 61.2 & 47.7\\
			    ours (L6 x V4 SH FT) & \textbf{52.0} & \textbf{40.7} \\

                \midrule
				% Pavllo et al.
				CVPR
				\citep{Pavllo2019}(CPN FT) & 46.8 & 36.5\\
				ours (L6 x V4 CPN FT) & \textbf{44.8} & \textbf{35.6}\\
				\midrule
				% Martinez et al.
				ICCV
				\citep{Martinez2017}(GT) & 45.5 & 37.1 \\
				% Hossain et al.
				ECCV
				\citep{Hossain2018}(GT) & 41.6 & 31.7\\
				% Lee et al.
				ECCV
				\citep{lee2018}(GT) & 38.4 & - \\
				% Zhao et al.
				CVPR
				\citep{Zhao2019}(GT) & 40.8 & 31.45\\
				% Pavllo et al.
				CVPR
				\citep{Pavllo2019}(GT) & 37.8 & 28.2 \\
				ours (L6 x V4 GT) & \textbf{33.4} & \textbf{26.1 }\\
				
				\bottomrule
			\end{tabular}
	\end{center}
	\caption{Performance impacted by 2D detectors under Protocol 1 and Protocol 2. {\bf (PT)} - pre-trained, {\bf (FT)} - fine-tuned, {\bf SH} - stacked hourglass.}
	\label{tb:tb4}
\end{table}

We investigated the impact of 2D pose detection on our 3D pose estimation performance by exploring several widely adopted 2D detectors. Firstly, we utilized the pre-trained \emph{Stacked Hourglass network} (SH) \citep{Newell2016} on the MPII dataset to extract 2D keypoint locations within the ground-truth bounding boxes. We also applied the results of fine-tuned SH model on the Human3.6M dataset developed by \citep{Martinez2017}. Researchers also investigated automated methods with detected bounding boxes for 2D human pose detection, such as \emph{Simple baselines for human pose estimation} \citep{xiao2018}, \emph{Deep high-resolution representation for human pose estimation} (HRnet) \citep{sun2019} or \emph{Cascaded Pyramid Network} (CPN) \citep{Chen2017} together with \emph{Mask R-CNN} \citep{He2017} and \emph{ResNet-101-FPN} \citep{Lin2017} as the backbone. We applied the pre-trained SH, fine-tuned SH, and fine-tuned CPN models \citep{Pavllo2019}  as the 2D detectors for performing a fair comparison, as shown in Table \ref{tb:tb4}.

The big difference between the pre-trained and fine-tuned models are the 2D human joints estimation accuracy and number of joints. Based on the results of our experiment, our network can learn different joint label information. MPII has 16 joints which missed the neck/nose joint in the Human3.6M dataset. Although COCO dataset has the same joint number, the order of the labels of joints is different from Human3.6M. To get a more accurate 3D joints position result, we utilize a fine-tuned model to get the corresponding 2D joints on Human3.6M. Furthermore, in the second part of Table \ref{tb:tb1}, we show the results with ground-truth (GT) 2D input. For both cases, our attention model demonstrates a clear advantage by utilizing the temporal information. 

\begin{figure}
  \includegraphics[width=\linewidth]{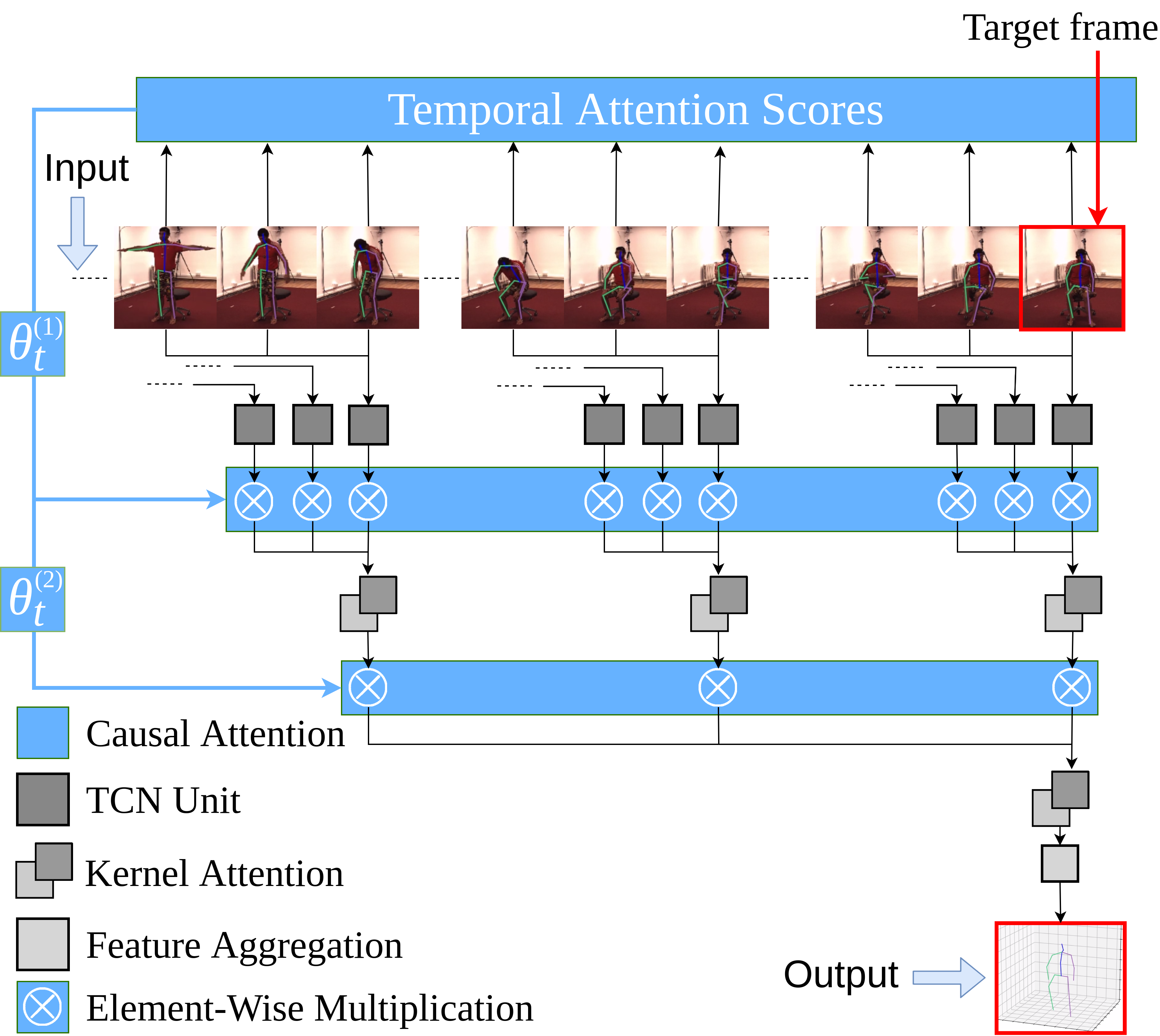}
  \caption{An example of a 4-layer architecture for causal attention-based temporal convolutional neural network.}\label{fig:causal}
\end{figure}

\subsubsection*{\bf Causal Attention Results}
\begin{figure}[]
\centering
\subfloat{\includegraphics[width= 2.5cm]{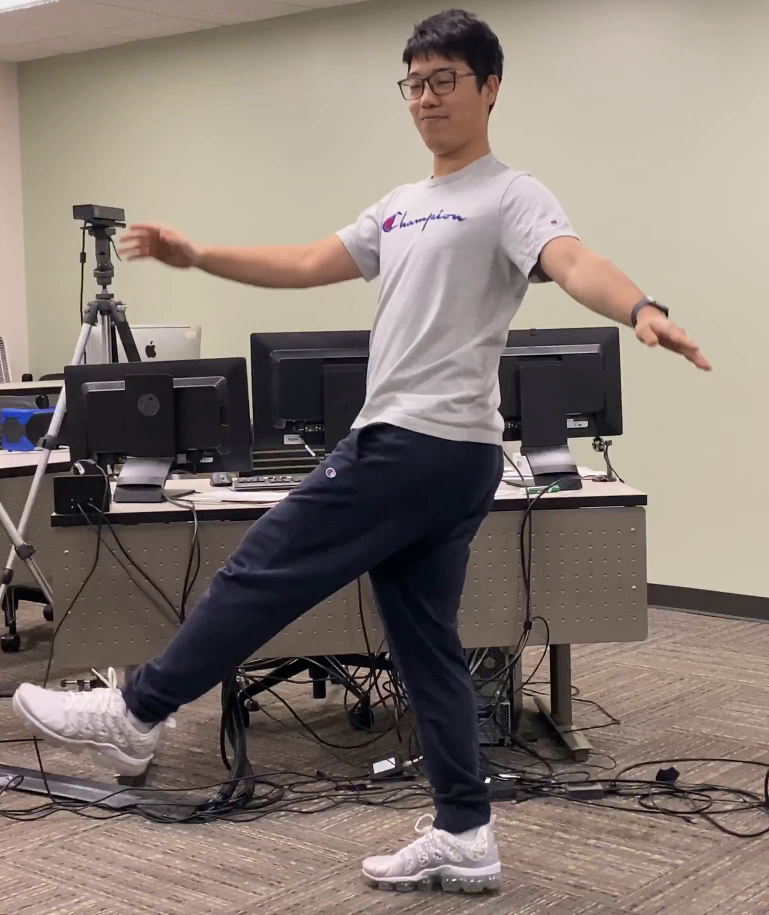}}
\subfloat{\includegraphics[width= 5.3 cm]{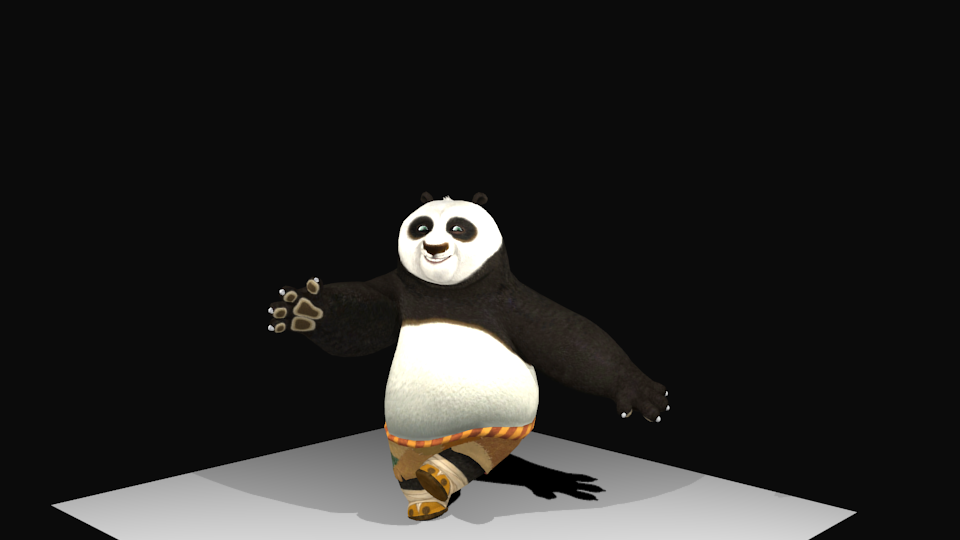}}\\
\subfloat{\includegraphics[width= 2.7 cm]{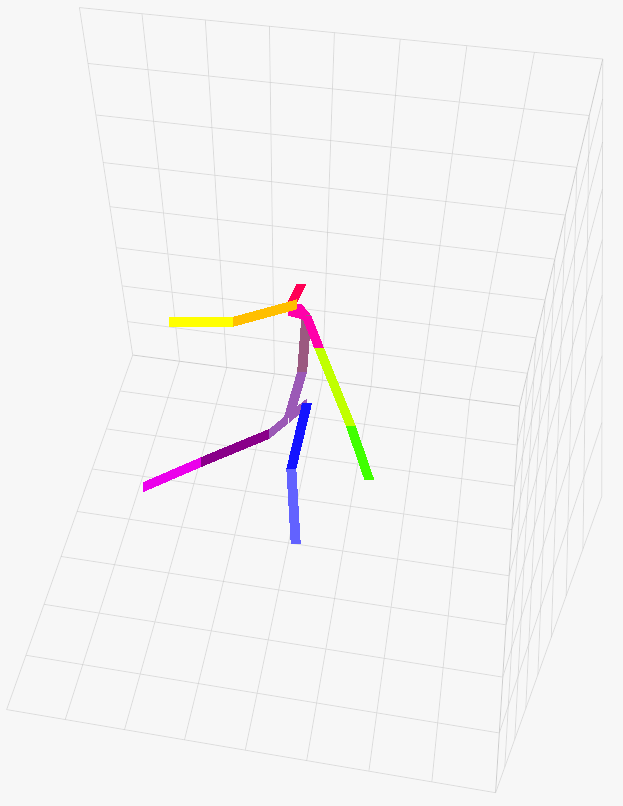}}
\subfloat{\includegraphics[width= 5.1cm]{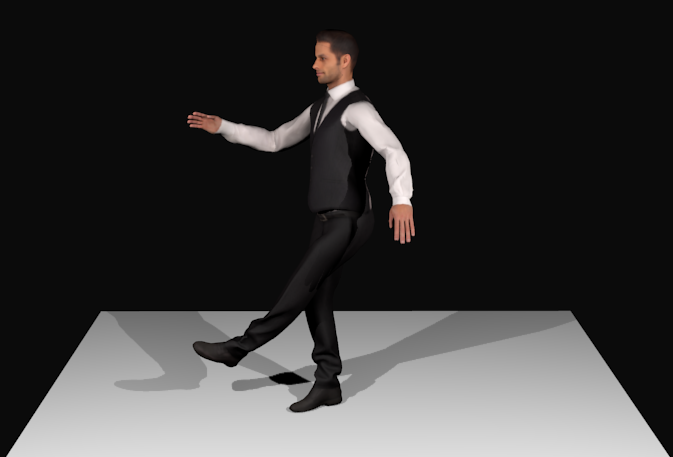}}
\caption{3D reconstruction results from different angles.}
\label{Multi-angle}
\end{figure}

To facilitate real-time performance for potential interactive applications, we also investigate a causal attention based network that estimates the target pose by only processing the current frame and its previous frames. The architecture of the causal attention model is shown in Fig. \ref{fig:causal}. The architecture is similar to the one described in Fig. \ref{fig:architecture}, but here we only consider the left half of the input video sequence. The number of input frames can also be determined by the number of layers of the model, but it shifts to the $\frac{N-1}{2}$ previous frames, where $N$ is the corresponding number of frames in the full-model illustrated in Fig. \ref{fig:architecture} . For example, for the configuration of  $L4 \times V2 \times N27$, $27$ causal frames are fed into the network (included the target frame); while $L5 \times V3 \times N81$ requires $81$ causal frames as the input. Similarly, to verify the performance, we implemented three different prototypes according to the number of layers and levels, as shown in Table \ref{tb:causal}. Horizontally, each row indicates a different prototype of the causal model. Vertically, each column indicates a different 2D detector. We provide a side-by-side comparison with the results in the recent CVPR paper on the same problem with various 2D detectors \citep{Pavllo2019}. Even our causal model only considers casual input frames compared to the TCN based semi-supervised approach in \citep{Pavllo2019}, the results of our method (ATCN + MDC) demonstrate higher accuracy consistently. In particular, more noticeable improvements are achieved as the number of input frames increases. The result of real-time processing using causal model is shown in Fig. \ref{Multi-angle}.

\begin{table}[h]
	\begin{center}
			\begin{tabular}{l|c|c|c|c}
				\toprule
				Method & SH PT  & SH FT & CPN FT & GT \\
				\citep{Pavllo2019} & - & - & 49.0 & -\\
				Ours(n=27) & - & 56.8 & 49.4 & 39.7\\
				Ours(n=81) & - & 55.7 & 47.5 & 37.1\\
				Ours(n=243) & 59.2 & 54.9 & 46.6 & 35.5\\
				\bottomrule
			\end{tabular}
	\end{center}
	\caption{Bottom-table: Causal sequence processing performance in terms of the different 2D detectors  under $Protocol \#1$. PT - pre-trained, FT - fine-tuned, GT - ground-truth, SH - stacked hourglass, CPN - cascaded pyramid network.}
	\label{tb:causal}
\end{table}

\begin{figure}[h!]
\subfloat{\includegraphics[width= \linewidth]{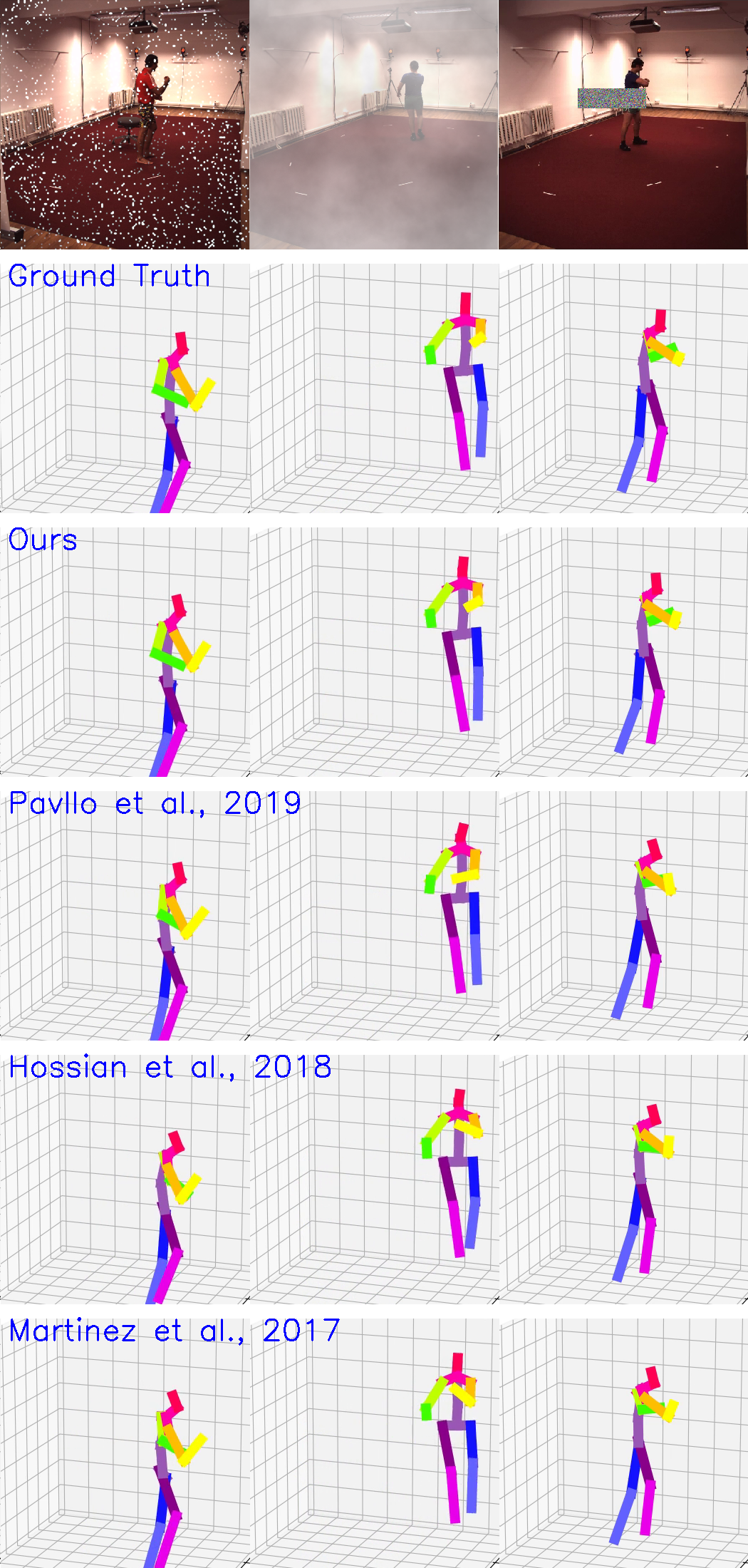}}\\
\caption{Samples of synthesized outdoor environment on the Human3.6M dataset and their 3D pose estimation.}\label{fig:noisyEffects}
\end{figure}

\begin{table*}[h!]
	\begin{center}
			\begin{tabular}{l|c c c c}
				\toprule
				 Noise & ICCV\citep{Martinez2017} & ECCV\citep{Hossain2018} & CVPR\citep{Pavllo2019} & Ours\\
				\midrule
				GT/GT & 37.1 & 31.7 & 28.1 & 26.1\\
				GT/GT + $\mathcal{N}(0,5)$  & 46.7 & 37.5 & 30.9 & 28.3\\
				GT/GT + $\mathcal{N}(0,10)$ & 52.8 & 49.4 & 39.3 & 36.7\\
				GT/GT + $\mathcal{N}(0,15)$ & 60.0 & 61.8 & 50.3 & 42.5\\
				GT/GT + $\mathcal{N}(0,20)$ & 70.2 & 73.7 & 62.2 & 56.2\\
				\bottomrule
			\end{tabular}
	\end{center}
	\caption{$Protocol \#2$ measurement on the estimation results from the simulated scenes. Training and testing on ground truth 2d joint locations plus different levels of additive gaussian noise.}
	\label{tb:noiseTable}
\end{table*}

\subsection{Performance on Videos in-the-wild}
\begin{figure}
    \centering
    \subfloat[Protocol 1: joint error analysis after adding noise across frames in Human3.6M \emph{Walking S9} left ankle.]{\includegraphics[width= \linewidth]{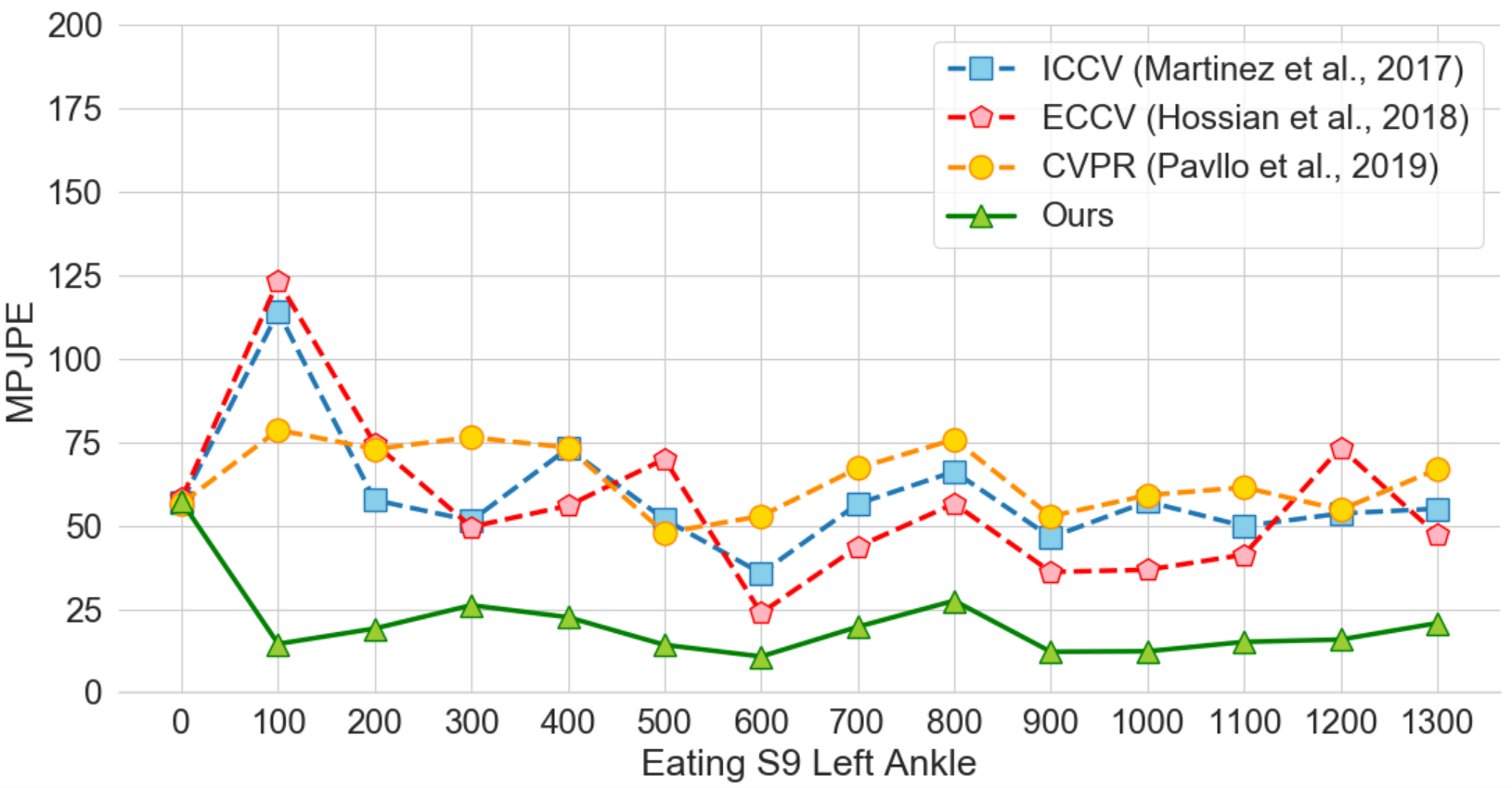}\label{fig:noise_joint1}}\\
    \subfloat[Protocol 1: joint error analysis after adding noise across frames in Human3.6M \emph{Smoking S9} left elbow. ]{\includegraphics[width= \linewidth]{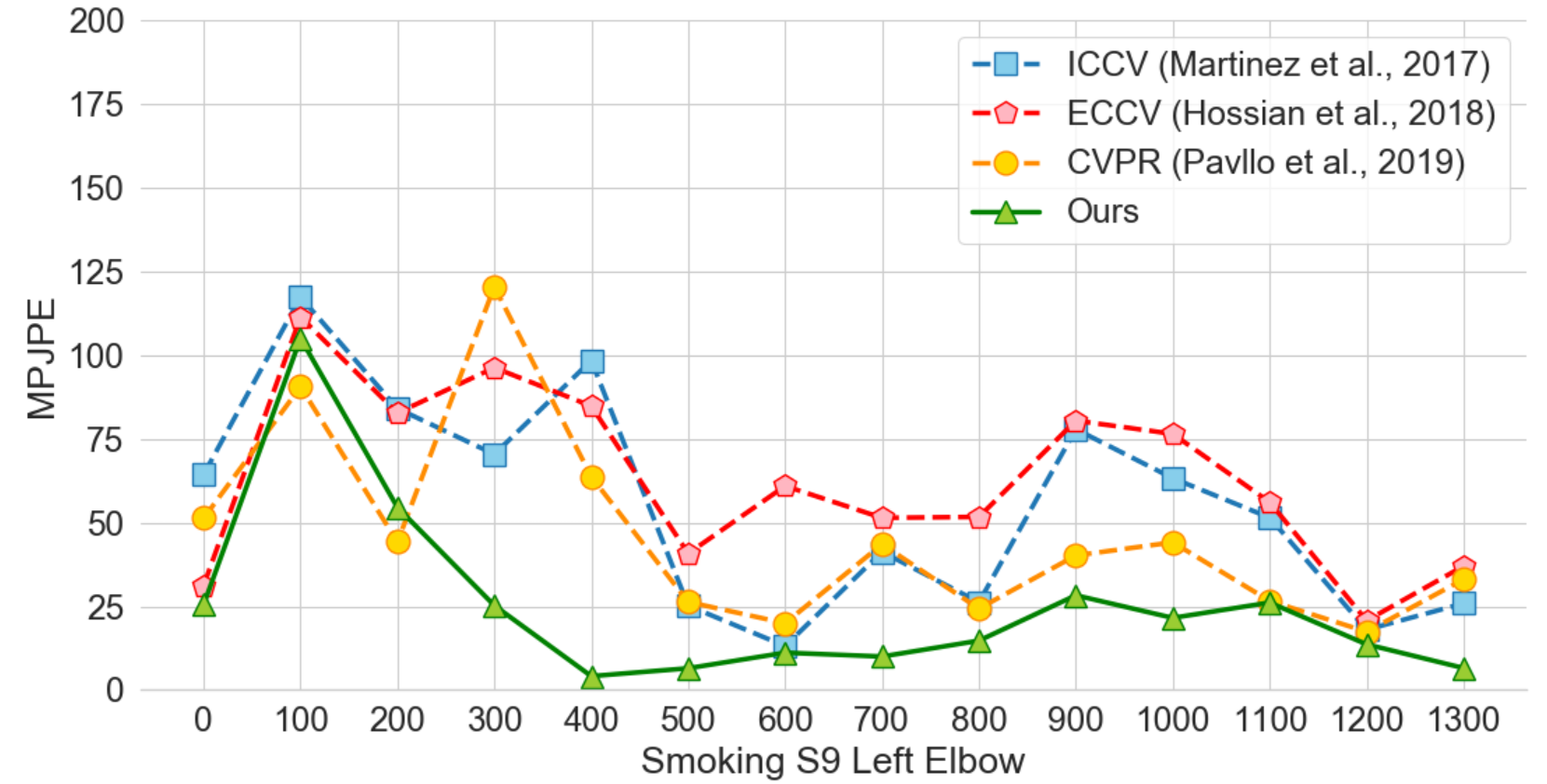}\label{fig:noise_joint2}}
    \caption{Joint-wise analysis and comparison on the outdoor simulated scenes. }\label{fig:noise_joint-wise-across-frames}
\end{figure}

\begin{figure*}[]
\subfloat{\includegraphics[width= \linewidth]{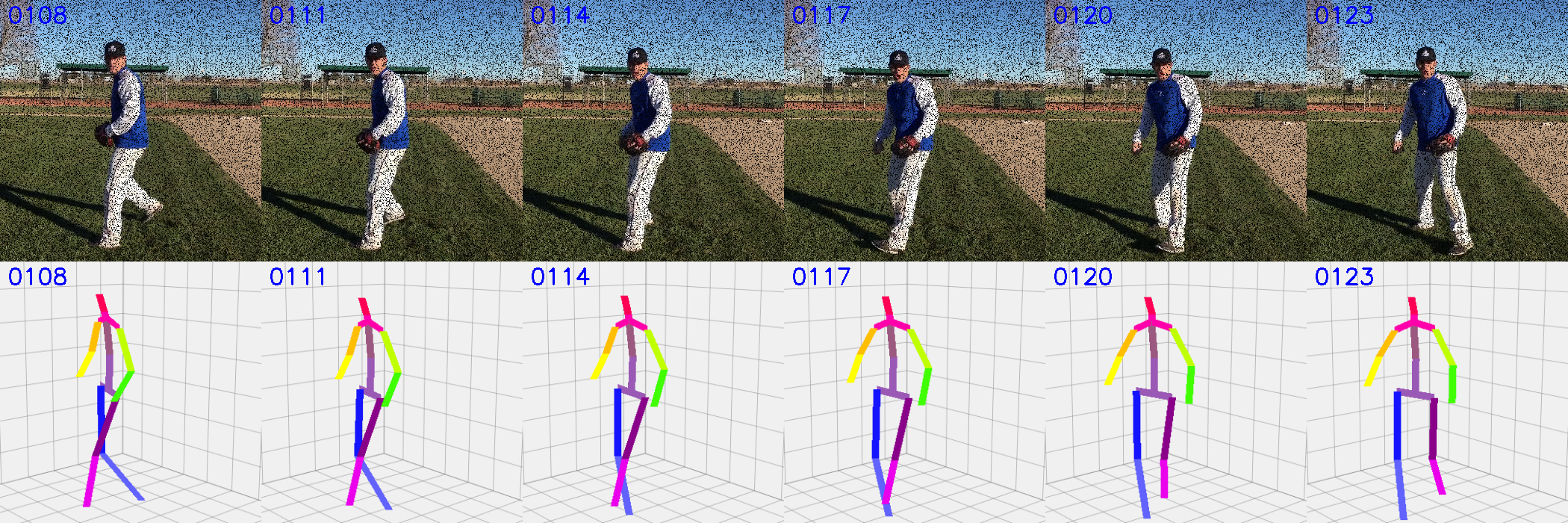}}\\
\subfloat{\includegraphics[width= \linewidth]{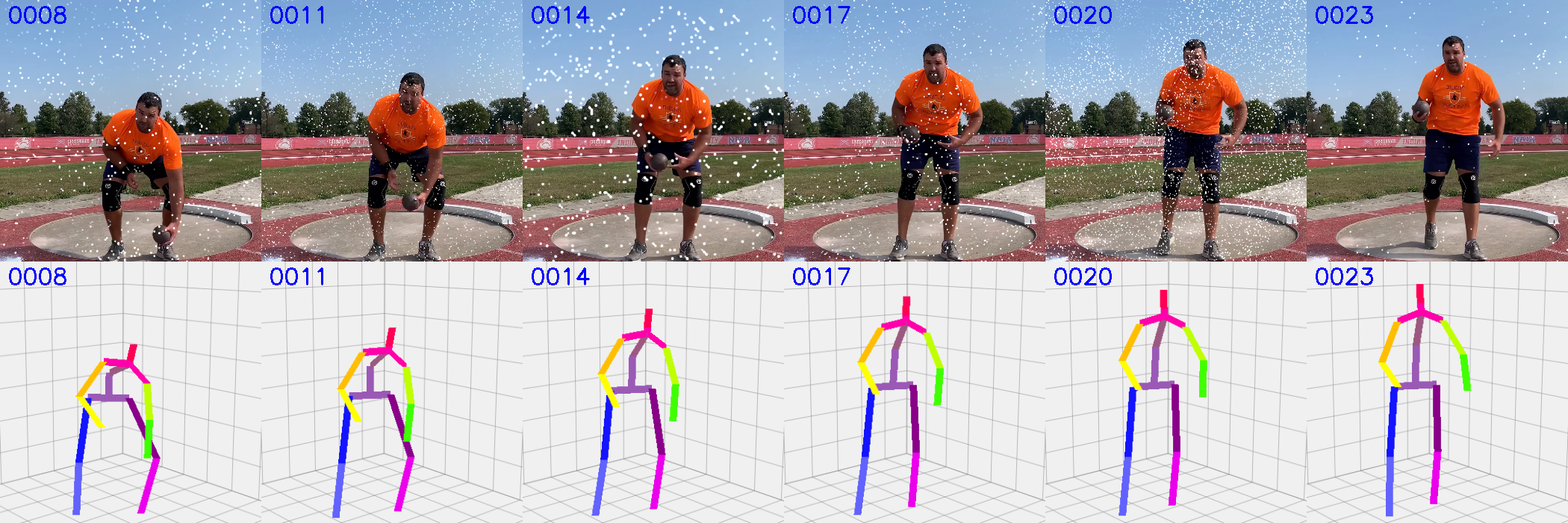}}\\
\subfloat{\includegraphics[width= \linewidth]{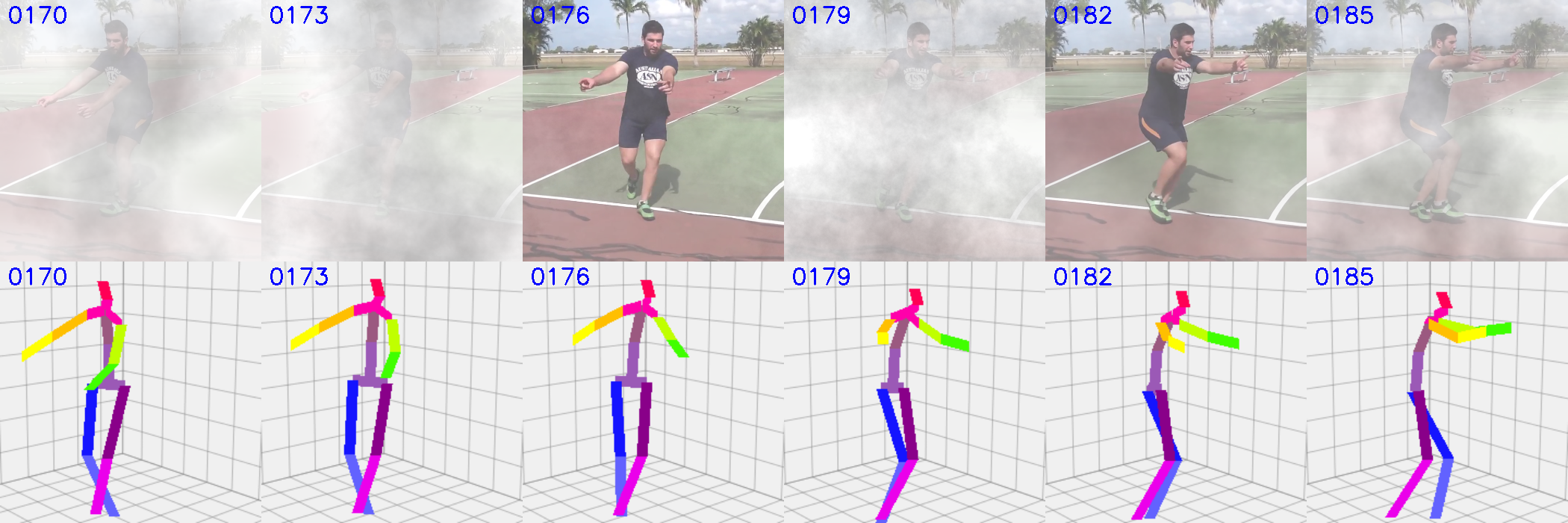}}
\caption{Qualitative results on gathered in the wild videos: original frame sequence with added noises and the recovered 3D poses.}
\label{fig:wildvideonoise}
\end{figure*}

To evaluate the performance on videos in-the-wild, we validated our approach on both public datasets and online videos with the former emphasizing quantitative validation while the later demonstrating qualitative performance. While there exists limited datasets with accurate 3D pose in the wild, we adopt some of the standard activities with outdoor scene simulation to quantitatively evaluate the performance and compare with other approaches. In contrast to static background and cameras capture setting, outdoor has more dynamic and unrestricted environment with frequent occlusion and high variation in background/foreground objects appearance. Fig. \ref{fig:noisyEffects} shows several outdoor simulations on the standard activities with snow, fog, and occlusion effects (each column). The corresponding pose estimation results by different approaches are shown in each of the following rows. Table \ref{tb:noiseTable} provides the quantitative measurement on their output. In a similar manner, joint-wise analysis is conducted on a selected joint from the Human3.6M scene with the generated noises. One can see our approach consistently yields less MPJPE over the frames as shown in Fig. \ref{fig:noise_joint-wise-across-frames}. To quantitatively demonstrate the robustness and efficacy, various videos in the wild are collected online and added with extra noises, e.g. snow or fog effect. Fig. \ref{fig:wildvideonoise} shows satisfactory results are achieved, given the additional noises. For example,  in the foggy scene (row 5 and 6), the target person is almost occluded by the thick fog. Thanks to the attention model that successfully extracts temporal information from neighbor frames, the full 3D pose is correctly recovered. 

\begin{figure*}[]
% \addtocounter{subfigure}{1}
\subfloat{\includegraphics[width= \linewidth]{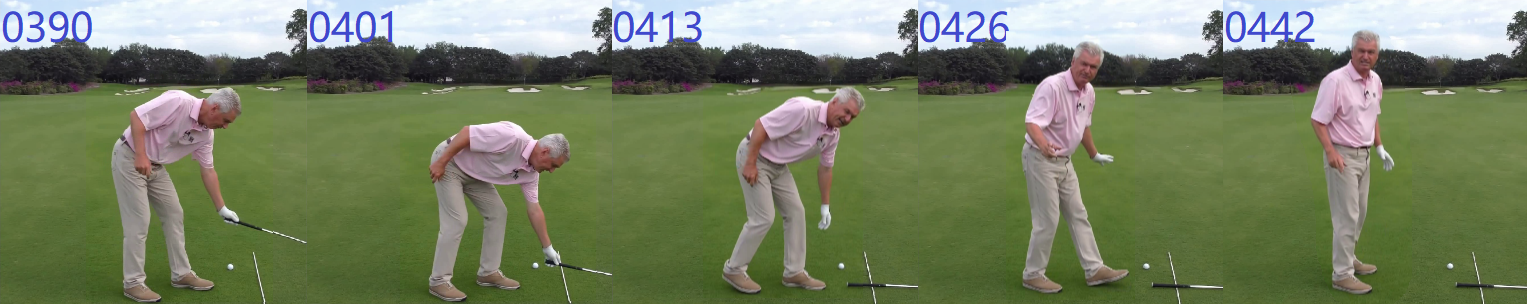}}\\
\subfloat{\includegraphics[width= \linewidth]{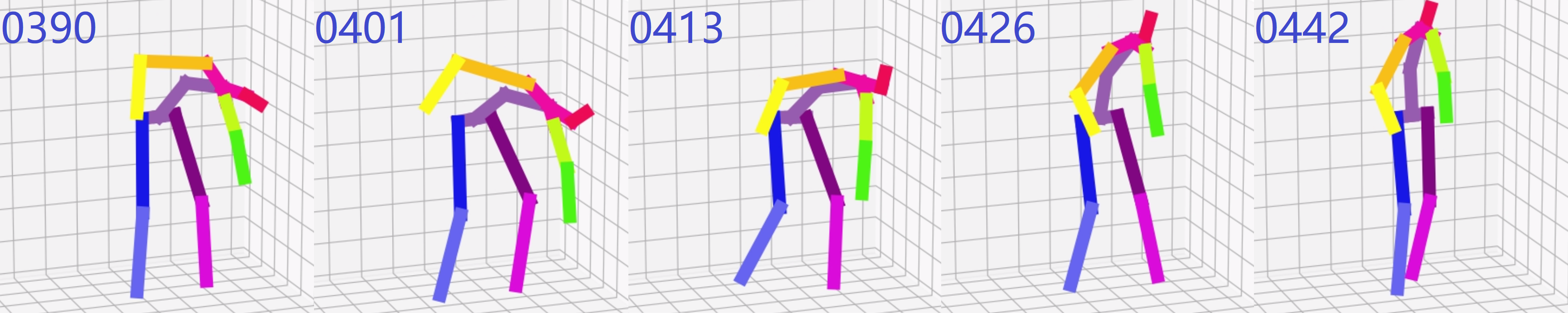}}\\
\subfloat{\includegraphics[width= \linewidth]{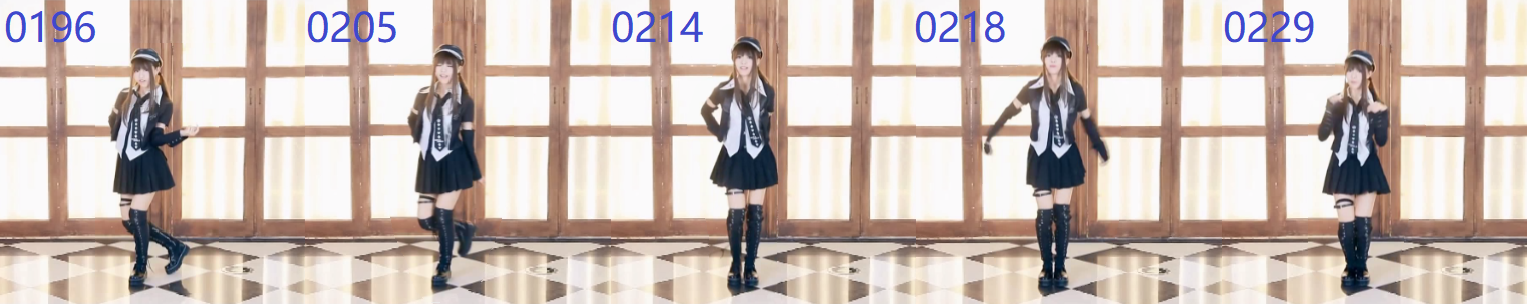}}\\
\subfloat{\includegraphics[width= \linewidth]{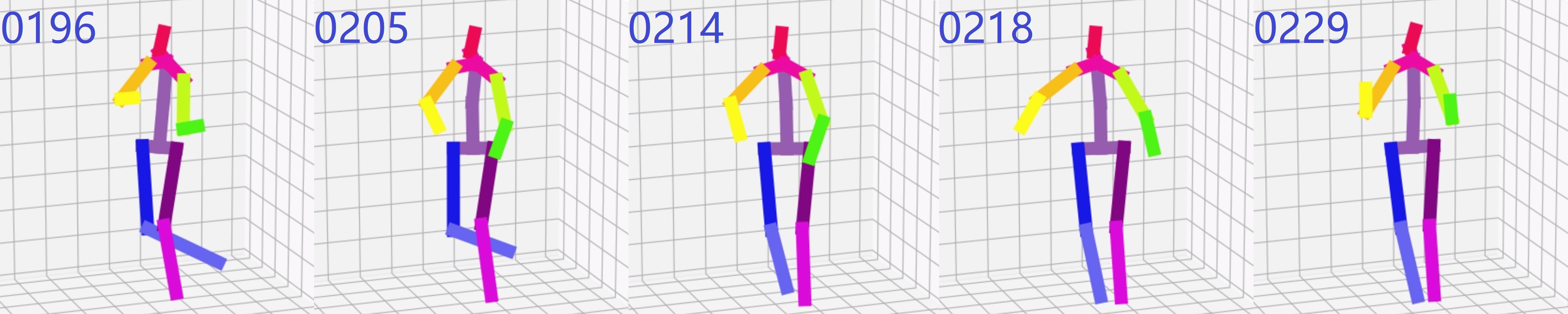}}\\
\subfloat{\includegraphics[width= \linewidth]{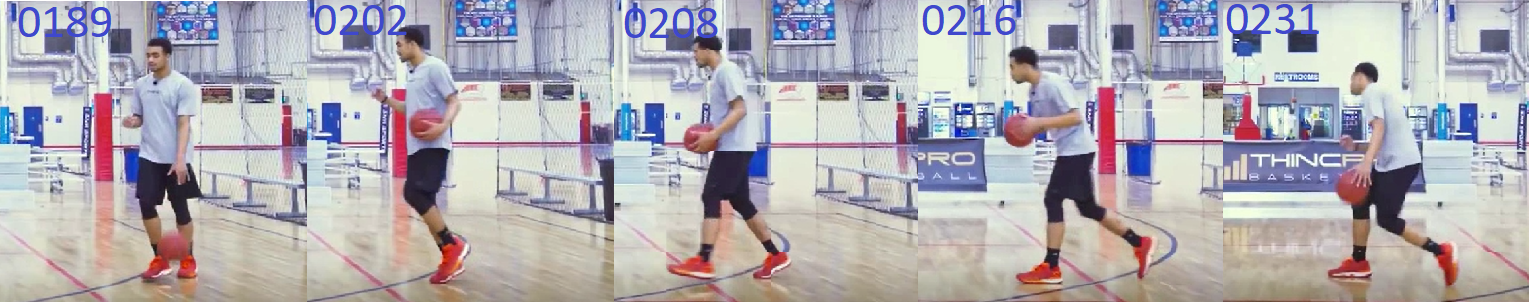}}\\
\subfloat{\includegraphics[width= \linewidth]{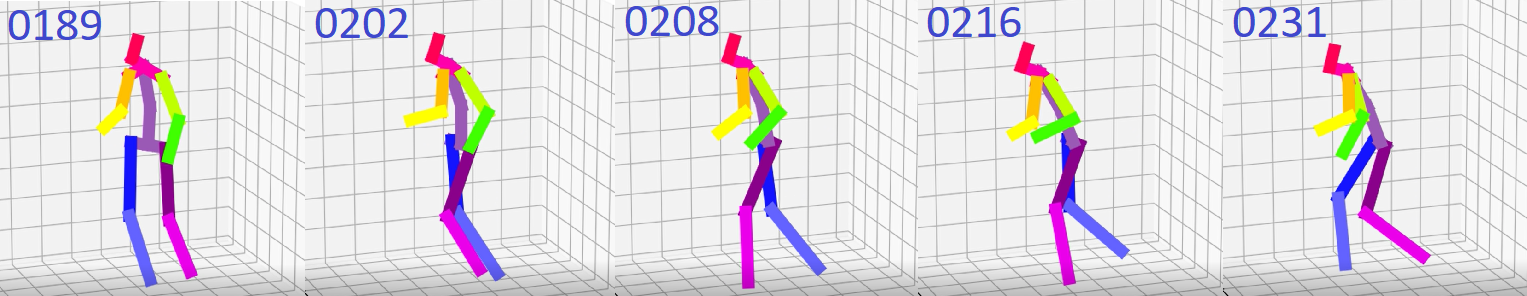}}\\
\caption{Qualitative results on gathered Youtube videos: original frame sequence and the recovered 3D poses.}

\label{fig:youtube}
\end{figure*}

To further demonstrate the temporal consistency, we gather online video sequences from YouTube and predict the 3D poses directly from these videos in the wild. Fig. \ref{fig:youtube} demonstrates the results of this experiment on various activities. Even though the input videos are either of low resolution or with fast motions, our approach is still able to estimate the 3D pose with satisfactory output. For example, for the dancing scenes (rows 1-2 and rows 9-10) and the skating scene (rows 5-6), given the presence of fast body movement and self-occlusion, the estimations are accurate enough to provide the corresponding 3D positions for each frame. To further verify the robustness, different sports activities with novel body poses (rows 3-4, rows 7-8, and rows 11-12) are processed. Our algorithm can faithfully capture and reproduce these pose details without requiring any additional offline training or manual preprocessing steps. In particular, for the challenging scene in rows 3-4, the target person has relatively casual dress with partial leg occlusion by the top costume The generated 3D pose from our attention model are visually plausible and resemble the user’s body motion very well.

\begin{figure}[]
\centering
\subfloat[Heavy occlusion.]{\includegraphics[width= 7.5cm]{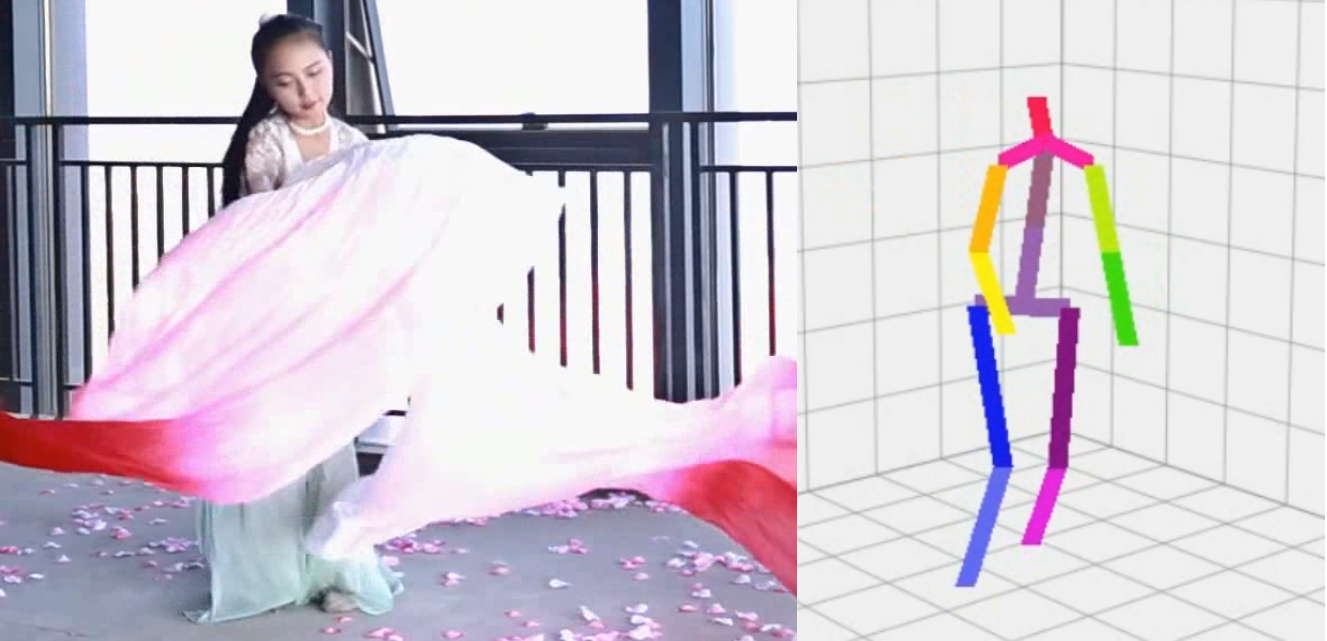}\label{heavy_oclusion}}\\
\subfloat[Instantaneous movement.]{\includegraphics[width= 7.5cm]{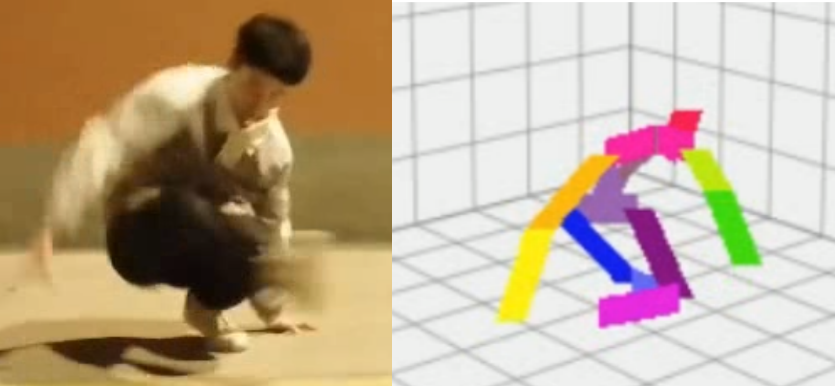}\label{fast_movement}}
\caption{Unresolved cases: there were a few failed frames from the tested wild videos, where severe occlusion and fast motion presented.}
\label{limitations}
\end{figure}

\section{Conclusion and Discussion}
In this paper, we present a novel and practical approach for 3D human pose estimation and reconstruction in unconstrained videos. In order to enhance temporal coherency, we integrate an attentional mechanism to the temporal convolutional neural network to guide the network towards learning informative representation. Moreover, we introduced a multi-scale dilation convolution, which is capable of capturing several levels of temporal receptive fields, achieving long-range dependencies among frames. Extensive experiments on benchmark  demonstrates that our approach improves upon the state-of-the-art and offers an effective, alternative framework to
address the 3D human pose estimation problem. The implementation is straightforward and can  adaptive corporate with standard convolution neural networks. For the input data, any off-the-shelf  2D  pose  estimation  systems,e.g. Mocap libraries, can be easily integrated in an ad-hoc fashion. 

Though our results outperform the state-of-the-art on public datasets, there are still some specific limitation remaining unresolved. Two examples are shown in Fig. \ref{limitations}. For example, when the performer's arms are crossing under the fans, it causes heavy occlusion with missing joints detection, thereby resulting in poor pose estimation, indicated in  Fig. \ref{limitations}\subref{heavy_oclusion}. In Fig. \ref{limitations}\subref{fast_movement}, when the leg has a very fast movement, our temporal system categorizes it as an outlier position rather than using them to contribute the pose inference. Another limitation is on  the inference accuracy for some multi-person human scenarios due to the limited training data on labeled multi-person 3D pose video datasets. However, if using the top-down 2D pose detecting algorithm with pose tracking, it would be possible to reconstruct multi-person 3D pose from a video. The tracking error may affect the temporal attention performance. Our future direction will explore a more generic framework that integrates the proposed attention model and person re-identification solution to handle instantaneous body part movements and heavy occlusions caused by multiple people.

\begin{figure*}[t!]
\subfloat{\includegraphics[width= \linewidth]{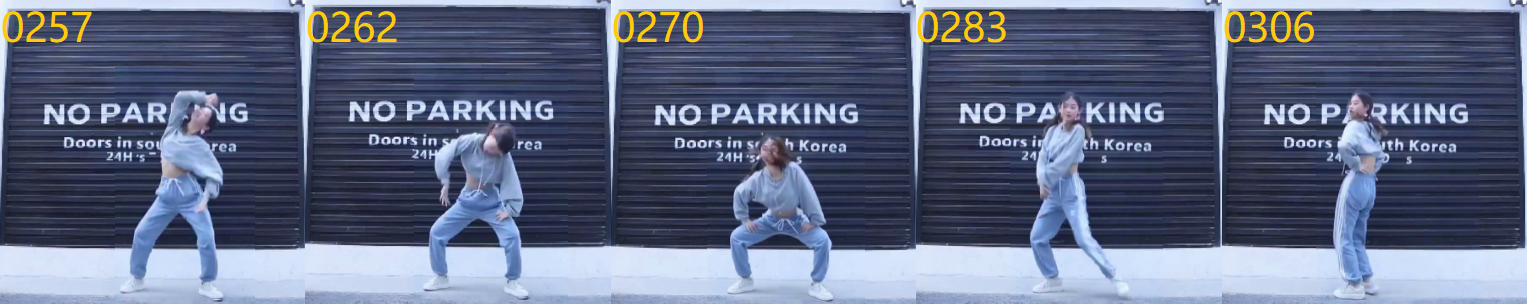}}\\
\subfloat{\includegraphics[width= \linewidth]{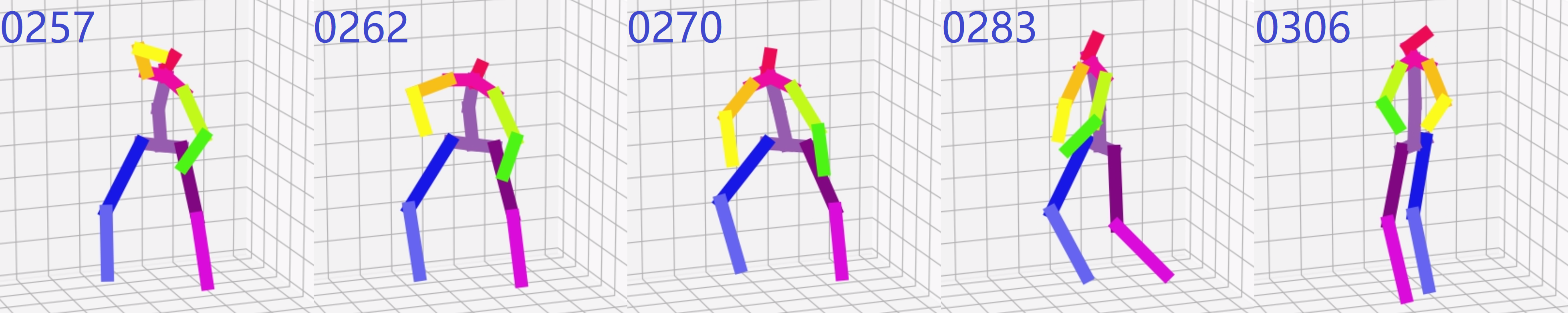}}\\
\subfloat{\includegraphics[width= \linewidth]{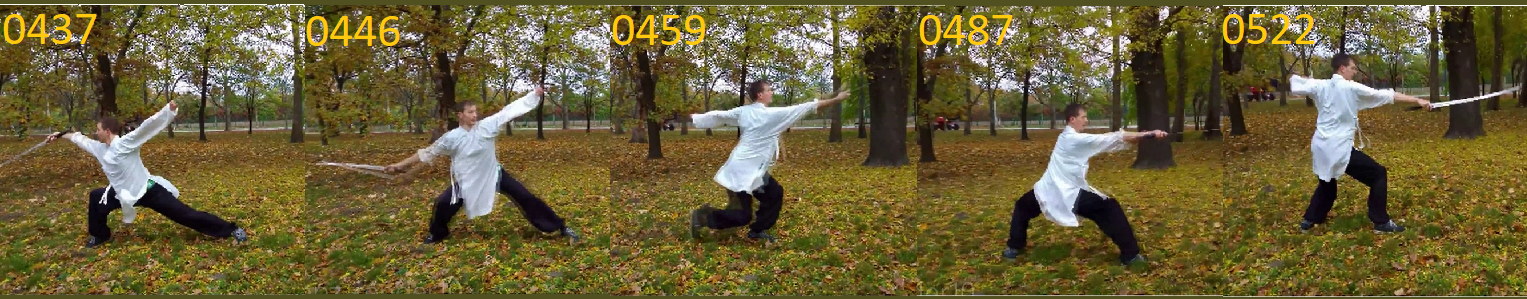}}\\
\subfloat{\includegraphics[width= \linewidth]{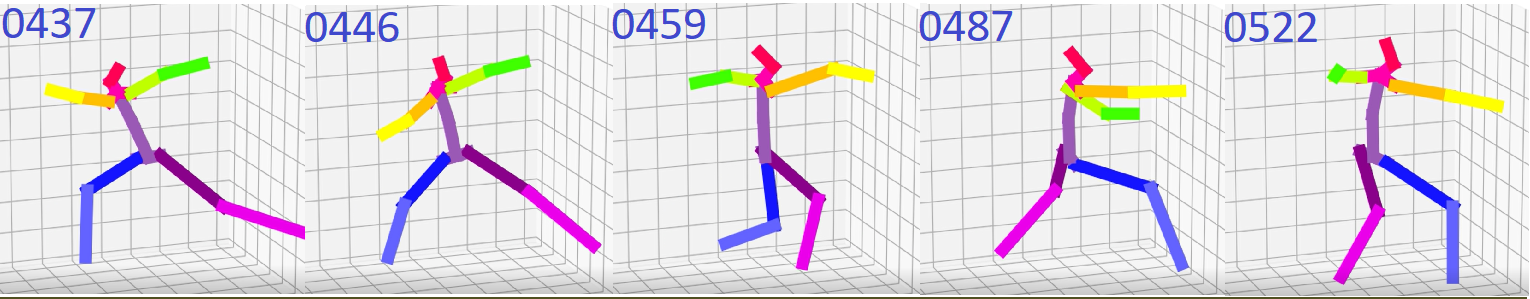}}\\
\subfloat{\includegraphics[width= \linewidth]{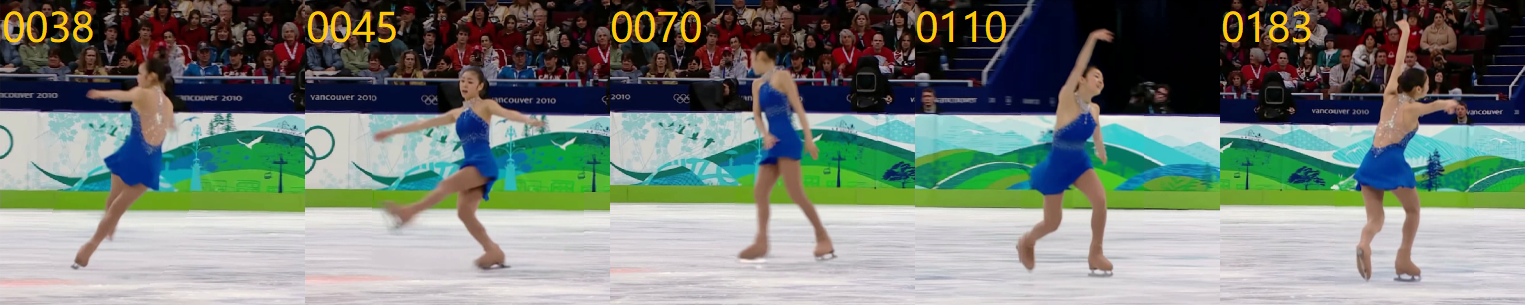}}\\
\subfloat{\includegraphics[width= \linewidth]{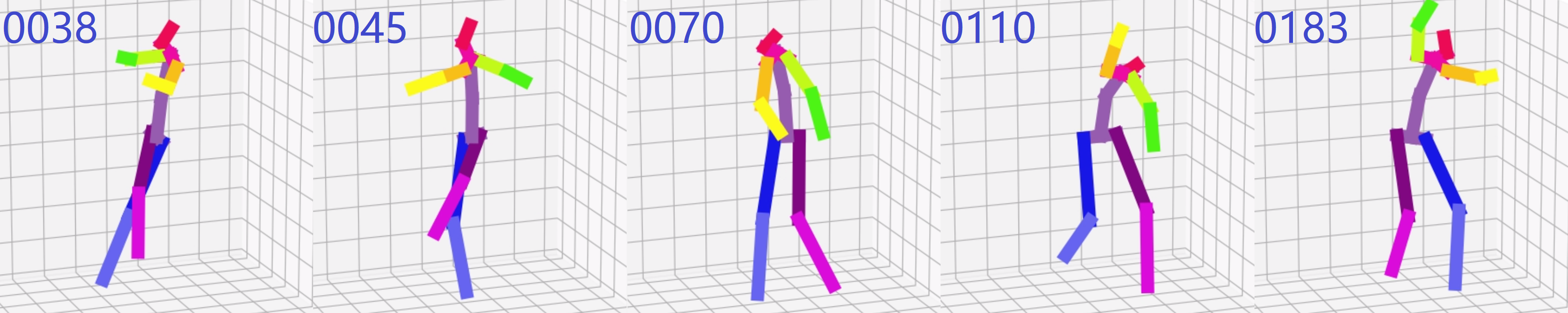}}
% \caption{Qualitative results on gathered Youtube videos: original frame sequence and the recovered 3D poses.}
\end{figure*}
% \addtocounter{figure}{-1} 

%\begin{acknowledgements}
%\end{acknowledgements}


\bibliographystyle{spbasic}      % basic style, author-year citations
% \bibliographystyle{spmpsci}      % mathematics and physical sciences
% \bibliographystyle{spphys}       % APS-like style for physics
%\clearpage
\bibliography{reference_file}

% \begin{thebibliography}{}
% %
% % and use \bibitem to create references. Consult the Instructions
% % for authors for reference list style.
% % Format for Journal Reference
% Author, Article title, Journal, Volume, page numbers (year)
% Author, Book title, page numbers. Publisher, place (year)
% \end{thebibliography}

\end{document}